\documentclass[10pt,twocolumn,letterpaper]{article}

\usepackage{cvpr}              
\definecolor{cvprblue}{rgb}{0.21,0.49,0.74}
\usepackage[pagebackref,breaklinks,colorlinks,allcolors=cvprblue]{hyperref}
\usepackage{algorithm}
\usepackage{algorithmic}
\usepackage{amsmath}
\usepackage{xcolor}
\usepackage{booktabs}
\usepackage{multirow}
\usepackage{amssymb}
\usepackage{bbding}
\usepackage{graphicx}
\definecolor{ggray}{RGB}{120,120,120}
\definecolor{tableblue}{RGB}{65,105,225}
\definecolor{tablered}{RGB}{220,20,60}
\usepackage{indentfirst}
\usepackage[normalem]{ulem}
\useunder{\uline}{\ul}{}
\usepackage[accsupp]{axessibility}
\usepackage[table]{xcolor}
\usepackage{cuted}
\usepackage{makecell}
\usepackage[accsupp]{axessibility}  

\newcommand{\boldparagraph}[1]{\noindent\textbf{#1}\ }

\def\paperID{8912} 
\def\confName{CVPR}
\def\confYear{2026}

\title{SPAN: Spatial-Projection Alignment for Monocular 3D Object Detection}

\author{Yifan Wang\textsuperscript{1},
Yian Zhao\textsuperscript{2},
Fanqi Pu\textsuperscript{1},
Xiaochen Yang\textsuperscript{3}, \\
Yang Tang\textsuperscript{4}\thanks{Corresponding authors.}, 
Xi Chen\textsuperscript{4},
Wenming Yang\textsuperscript{1}\footnotemark[1], \\
\textsuperscript{1}Shenzhen International Graduate School, Tsinghua University \\
\textsuperscript{2}School of Electronic and Computer Engineering, Peking University \\
\textsuperscript{3}School of Mathematics and Statistics, University of Glasgow \\
\textsuperscript{4}Basic Algorithm Center, PCG, Tencent \\
\texttt{\small yf-wang23@mails.tsinghua.edu.cn, ethanntang@tencent.com, yang.wenming@sz.tsinghua.edu.cn} \\
}

\begin{document}
\maketitle
\begin{abstract}
\noindent
Existing monocular 3D detectors typically tame the pronounced nonlinear regression of 3D bounding box through decoupled prediction paradigm, which employs multiple branches to estimate geometric center, depth, dimensions, and rotation angle separately.
Although this decoupling strategy simplifies the learning process, it inherently ignores the geometric collaborative constraints between different attributes, resulting in the lack of geometric consistency prior, thereby leading to suboptimal performance.
To address this issue, we propose novel \textbf{S}patial-\textbf{P}rojection \textbf{A}lig\textbf{n}ment~(\textbf{SPAN}) with two pivotal components:
(i). \textbf{\textit{Spatial Point Alignment}} enforces an explicit global spatial constraint between the predicted and ground‑truth 3D bounding boxes, thereby rectifying spatial drift caused by decoupled attribute regression.
(ii). \textbf{\textit{3D-2D Projection Alignment}} ensures that the projected 3D box is aligned tightly within its corresponding 2D detection bounding box on the image plane, mitigating projection misalignment overlooked in previous works.
To ensure training stability, we further introduce a \textbf{\textit{Hierarchical Task Learning}} strategy that progressively incorporates spatial-projection alignment as 3D attribute predictions refine, preventing early stage error propagation across attributes.
Extensive experiments demonstrate that the proposed method can be easily integrated into any established monocular 3D detector and delivers significant performance improvements.
Project Page: \url{https://wyfdut.github.io/SPAN/}
\end{abstract}
    
\section{Introduction}
\label{sec:intro}

\begin{figure}[t]
    \centering
    \includegraphics[width=0.95\linewidth]{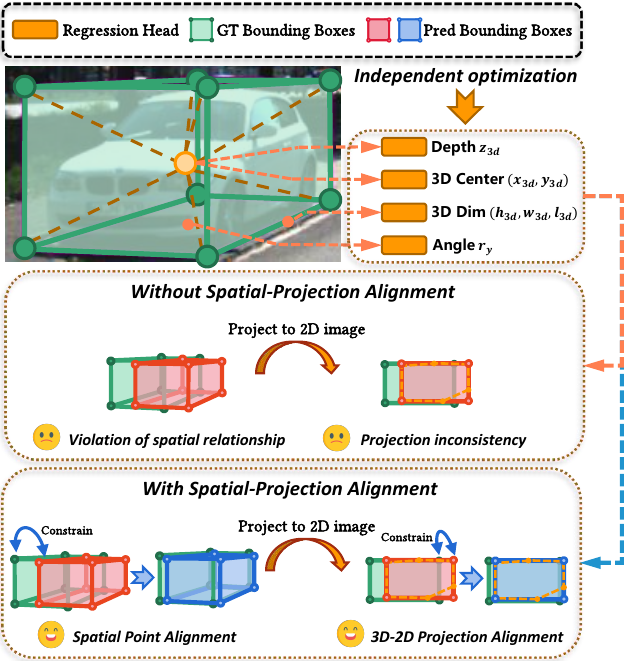}%
    \vspace{-0pt}
    \caption{
    Previous methods typically neglect the geometric collaborative constraints between different attributes, resulting in spatial errors and projection misalignment. Our method aligns 3D corners and matches 3D projections with 2D boxes to improve detection accuracy and consistency.}
    \label{fig:intro}
    \vspace{-5pt}
 \end{figure}

 3D object detection underpins both autonomous driving and robotic perception by delivering essential spatial awareness of the surrounding environment. Compared with LiDAR~\cite{PV_RCNN, PV_RCNN++, wu2023virtual, fshnet} or stereo~\cite{dsgn++, side, wu2023semi, dsc3d, li2024real} based counterparts, monocular 3D detection infers full spatial information from a single RGB camera, offering clear advantages in cost efficiency and deployment flexibility.

The attributes of a 3D bounding box are typically represented as $\left( {{x_{3d}},{y_{3d}},{z_{3d}},{h_{3d}},{w_{3d}},{l_{3d}},{r_y}} \right)$. As shown in Fig.~\ref{fig:intro}, most existing monocular 3D detection frameworks \cite{gupnet, monodde, monodetr, monocd, fd3d, gupnet++, monodgp} employ a decoupled regression paradigm, wherein seven degrees‑of‑freedom parameters are predicted by separate heads. 
Although this factorization streamlines learning objectives, its intrinsic disregard for geometric collaborative constraints has gradually become a major bottleneck that impedes further performance gains.

Predicting the individual attributes of a 3D bounding box in isolation often violates the inherent spatial relationships among those attributes.
Because these attributes lack coordinated consistent constraints, the resulting predictions cannot fully align with the ground-truth cuboids in 3D space, thereby degrading localization accuracy.
Some studies have sought to incorporate geometric constraints to address these issues.
Early attempts such as \cite{deep3dbox, shift_rcnn, rtm3d} solved an overdetermined system of equations to recover depth. However, even slight perturbations in the 2D box can induce large fluctuations in the solution space.
More recent studies introduce richer geometric priors and data-augmentation strategies, yet they do not explicitly model spatial or projection constraints during training.
For instance,~\cite{homography} leverages a global homography assumption to balance different objectives, but it lacks fine-grained correction for nonplanar targets or local shape variations.
Augmentation schemes like 3D Copy-Paste~\cite{3d_copy_paste} and MonoPlace3D~\cite{monoplace3d} enrich sample diversity using geometric priors, yet they perform no rigorous verification of 3D-2D projection consistency and may introduce implausible occlusions.
MonoDGP~\cite{monodgp} incorporates geometric error priors into projection formulas to mitigate depth biases, but still regresses attributes independently and thus overlooks the need for unified consistency across parameters.
Collectively, geometric collaborative constraints have yet to be adequately exploited in prior monocular 3D detection frameworks, which has limited the utilization of monocular cues.

Our work addresses this gap by explicitly integrating \textit{Spatial-Projection Alignment} into an end-to-end framework.
To be concrete as shown in Fig.~\ref{fig:intro}, our central idea is to impose geometrically collaborative constraints on every set of 3D bounding box attributes and to optimize them jointly to enhance both spatial and projection coherence.
Specifically, we first propose \textit{Spatial Point Alignment}. In contrast to ROI-10D \cite{roi-10d} and MonoDIS \cite{simonelli2019disentangling} that simply regress corners as an auxiliary task, our loss is applied to the eight corner coordinates derived from the primary 7-DoF attributes. 
This directly regularizes the main box parameters by constraining them to align with the ground-truth corners, thereby enforcing geometric integrity.
Second, we propose \textit{3D-2D Projection Alignment}. Unlike analytical solvers like Deep3DBox \cite{deep3dbox} that are sensitive to 2D detection noise, our differentiable approach provides a robust, gradient-based learning signal. 
It projects the corners of each predicted 3D box onto the image plane and constrains the minimal enclosing rectangle of this projection to align with its corresponding 2D detection box.

During early training, the predicted 3D bounding box parameters are subject to substantial noise, rendering our geometric collaborative constraints unreliable.
Consequently, after introducing the Spatial-Projection Alignment, we observed marked instability at the outset of model training.
To mitigate this, we adapt and refine GUPNet's~\cite{gupnet} \textit{Hierarchical Task Learning} strategy. 
Briefly, this scheme assigns greater weight to the 2D and 3D attribute regression objectives during the initial training epochs, and gradually increases the emphasis on the more challenging geometric collaborative constraints as learning progresses, preserving training stability in the early phase.

In summary, our contributions are listed as follows:
\begin{itemize}
    \item We observe that the prevailing decoupled regression paradigm in monocular 3D detection often overlooks the inherent spatial and projection relationships among the bounding-box attributes. This gap can result in predictions that violate physical projection constraints, thereby limiting localization accuracy.
    \item We propose a unified geometric consistency optimization paradigm for monocular 3D object detection that comprises Spatial Point Alignment and 3D-2D Projection Alignment to ensure the predicted bounding box aligns spatially with the ground-truth cuboid and simultaneously satisfies the physical projection constraint between the 3D structure and its 2D bounding box. Both are integrated via a Hierarchical Task Learning strategy to ensure stable training. Our method innovatively combines alignment constraints with a phased scheduling mechanism, which has not been considered in previous work.
    \item The proposed Spatial-Projection Alignment can be employed as a plug-and-play module within mainstream monocular 3D detectors. Experimental results reveal that it elevates the accuracy of state-of-the-art monocular 3D detectors without introducing additional modules.
\end{itemize}

\begin{figure*}
   \centering
   \includegraphics[width=0.95\linewidth]{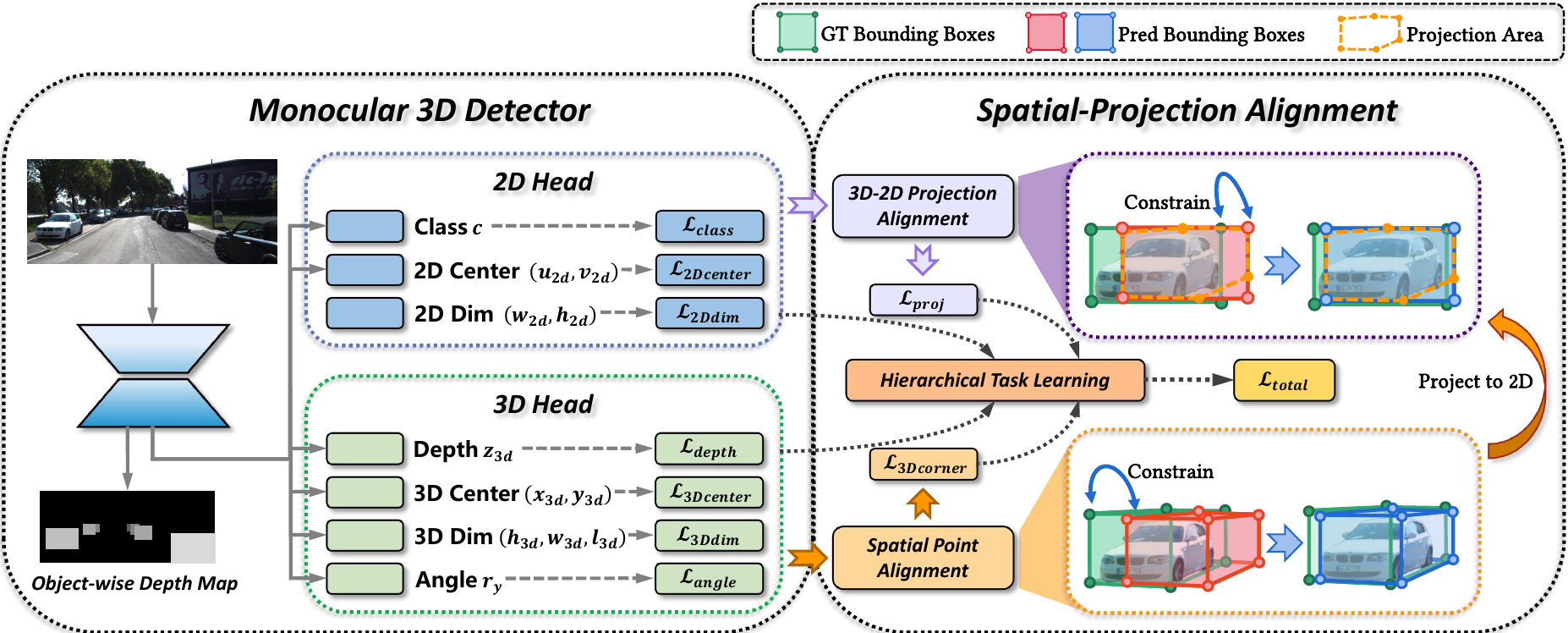}
   \vspace{-0pt}
   \caption{\textbf{Overview of the proposed Spatial-Projection Alignment (SPAN).} After a monocular detector predicts 2D and 3D attributes, SPAN adds two constrains: (a) Spatial Point Alignment that aligns predicted box corners with ground-truth in 3D space, and (b) 3D-2D Projection Alignment that constrains the projected 3D box to fit tightly inside the 2D detection. Hierarchical Task Learning controls the relative weights of these losses during training.}
   \vspace{-0pt}
   \label{fig:overview}
\end{figure*}
\section{Related Work}
\label{sec:related work}

\boldparagraph{Monocular 3D Object Detection.}
Monocular 3D object detection seeks to recover 3D bounding boxes from a single image.
However, the absence of direct depth cues renders this an ill-posed problem.
Early works such as Deep3DBox~\cite{deep3dbox} and Shift R-CNN~\cite{shift_rcnn} leverage 2D to 3D perspective geometry constraints to locate the 3D box center.
More recent approaches \cite{gupnet, monocon, monodde, monodetr, monouni, monocd, fd3d, monodgp} combine data-driven frameworks with learnable geometric priors.
For example, \cite{gupnet, deviant, gupnet++} introduce uncertainty into geometric projection-based depth estimation to improve detection accuracy.
\cite{monoasrh} employs a hybrid feature aggregation technique~\cite{rt-detr, rt-detrv4} to enhance the receptive field.
~MonoDETR~\cite{monodetr} incorporates an object-level foreground depth map and fuses regressed depth, geometric depth, and foreground depth to obtain a final depth estimate.
FD3D~\cite{fd3d} further supervises intermediate features using the foreground depth map to provide clearer guidance for foreground objects.
MonoCD~\cite{monocd} systematically analyzes the complementarity of diverse depth cues in geometric space and fuses them to mitigate positive and negative errors. 
MonoDGP~\cite{monodgp} studies geometric error in projected depth and proposes a perspective invariant correction formula.
Although these methods have achieved notable progress, the produced 3D boxes still exhibit geometric misalignment because each parameter is optimized independently, lacking spatial coherence. Our geometric collaborative loss alleviates this issue by imposing unified geometric constraints during end-to-end training.

\boldparagraph{Geometric Constraints in Mono3D.}
Geometric constraints provide crucial priors to alleviate monocular depth ambiguity.
Early works established foundational principles such as the projection constraint in Deep3DBox~\cite{deep3dbox}, which links 3D boxes to 2D detections, and inspired subsequent refinements in Shift R-CNN~\cite{shift_rcnn} and RTM3D~\cite{rtm3d}. Subsequent research has extended geometric constraint strategies in various ways.
Li \emph{et al}.~\cite{li2021monocular} recast the non-linear optimization in projection space as a differentiable geometric reasoning module.
Homography Loss~\cite{homography} exploits global geometric information to balance spatial relationships across different objects.
Lian \emph{et al}.~\cite{lian2022exploring} propose geometry-aware image and instance augmentation that encourages depth predictions to remain invariant under geometric transformations.
GUPNet++~\cite{gupnet++} refines the depth projection formula by propagating geometric uncertainty. 
Other works \cite{3d_copy_paste, monoplace3d} synthesize realistic 3D data using geometric priors to boost detector performance.
Nevertheless, most methods still underexploit the full potential of spatial and projection consistency, and their reliance on exact geometry often causes instability during early training. Our proposed SPAN integrates a Hierarchical Task Learning scheme that dynamically adjusts loss weights, ensuring stable training.

\section{Methodology}

\subsection{Overview}

As illustrated in Fig. \ref{fig:overview}, the proposed SPAN can be seamlessly plugged into the training pipeline of any monocular 3D detector. First, the detector’s original heads regress both 2D attributes and the primary 3D attributes. Next, it computes the eight corner points of the predicted 3D box and applies a \textit{Spatial Point Alignment} loss to constrain spatial consistency with the ground-truth corners. These corners are then projected onto the 2D plane, and a \textit{3D-2D Projection Alignment} loss is used to ensure that the horizontally aligned minimal enclosing rectangle of the projected points tightly coincides with the 2D detection box, thereby satisfying perspective constraints. Finally, the two geometric collaborative losses are dynamically weighted together with the regression losses of the 2D and 3D heads via \textit{Hierarchical Task Learning}, ensuring stable initial training and enhanced geometric refinement.

\subsection{Spatial Point Alignment}
\label{Spatial Point Alignment}

The regression head of a monocular 3D detector yields the 3D center $\left( {{x_{3d}},{y_{3d}}} \right)$, depth ${z_{3d}}$, 3D box dimensions $\left( {{h_{3d}},{w_{3d}},{l_{3d}}} \right)$, and the yaw angle $r_y$ around the $Y$ axis. Leaving the resulting 3D box center in the camera coordinate system as $\mathbf{C}=[x_{3d},y_{3d},z_{3d}]^{T}$, the eight corners of the cuboid $\{\mathbf{P}_i=(x_i,y_i,z_i)\}_{i=1}^8$ are then obtained as:
\begin{equation}
\begin{gathered}
    \mathbf{P}_i=\mathbf{C}+\mathbf{R}(r_y)\mathbf{P}_{l,[i]}, \\
    \mathbf{P}_l=\mathbf{D}_l
    \left[\begin{array}{@{} r r r r r r r r @{}}
    \frac{1}{2} & \frac{1}{2} & -\frac{1}{2} & -\frac{1}{2} & \frac{1}{2} & \frac{1}{2} & -\frac{1}{2} & -\frac{1}{2} \\[0.6ex]
    \frac{1}{2} & \frac{1}{2} & \frac{1}{2} & \frac{1}{2} & -\frac{1}{2} & -\frac{1}{2} & -\frac{1}{2} & -\frac{1}{2} \\[0.6ex]
    \frac{1}{2} & -\frac{1}{2} & -\frac{1}{2} & \frac{1}{2} & \frac{1}{2} & -\frac{1}{2} & -\frac{1}{2} & \frac{1}{2}
    \end{array}\right], \\
    \mathbf{D}_l=
    \begin{bmatrix}
    l_{3d} & 0 & 0 \\
    0 & h_{3d} & 0 \\
    0 & 0 & w_{3d}
    \end{bmatrix},\mathbf{R}\left(r_{y}\right)=
    \begin{bmatrix}
    \cos {r_y} & 0 & \sin {r_y} \\
    0 & 1 & 0 \\
    -\sin {r_y} & 0 & \cos {r_y}
    \end{bmatrix}, \\[0.6ex]
    \end{gathered}
    \label{p_i}
\end{equation}
where $\mathbf{R}\left(r_{y}\right)$ denotes the rotation matrix about the $Y$‑axis.

Having obtained the eight predicted vertices $\{\mathbf{P}_i\}_{i=1}^8$ and the eight corresponding ground-truth vertices $\{\mathbf{G}_i\}_{i=1}^8$ of each 3D bounding box, one might consider using the 3D GIoU to align these points.  However, directly computing the GIoU between two arbitrarily oriented 3D cuboids is computationally intensive, since it involves exact intersection calculations of convex polyhedra.  To sidestep this complexity, we employ the MGIoU~\cite{mgiou} scheme, which breaks the entire 3D overlap problem down into three simpler one‑dimensional projection problems.

Specifically, we begin by computing the three unique face normal vectors of the 3D bounding box:
\begin{equation}
    \mathcal{A}=\{\mathbf{a}_1,\mathbf{a}_2,\mathbf{a}_3\},\|\mathbf{a}_k\|=1;
    \label{face_normal}
\end{equation}
for each normal vector ${\mathbf{a}_k} \in \mathcal{A}$, we compute the projection of every vertex onto that vector as follows:
\begin{equation}
    \begin{gathered}
    \mathrm{proj}^{\left(i\right)}_{\mathbf{P},\mathbf{a}_k}=\mathbf{P}_i\cdot\mathbf{a}_k, \\
    \mathrm{~proj}^{\left(i\right)}_{\mathbf{G},\mathbf{a}_k}=\mathbf{G}_i\cdot\mathbf{a}_k.
    \end{gathered}
    \label{proj_vec}
\end{equation}
Next, we extract the interval endpoints from these projected values:
\begin{equation}
    \begin{gathered}
    {\min }_{{\mathbf{P}},{{\mathbf{a}}_k}} = \min_{1\leq i\leq8} ({\text{pro}}{{\text{j}}^{\left(i\right)}_{{\mathbf{P}},{{\mathbf{a}}_k}}}),{\max }_{{\mathbf{P}},{{\mathbf{a}}_k}} = \max_{1\leq i\leq8} ({\text{pro}}{{\text{j}}^{\left(i\right)}_{{\mathbf{P}},{{\mathbf{a}}_k}}}), \\
    {\min }_{{\mathbf{G}},{{\mathbf{a}}_k}} = \min_{1\leq i\leq8} ({\text{pro}}{{\text{j}}^{\left(i\right)}_{{\mathbf{G}},{{\mathbf{a}}_k}}}),{\max }_{{\mathbf{G}},{{\mathbf{a}}_k}} = \max_{1\leq i\leq8} ({\text{pro}}{{\text{j}}^{\left(i\right)}_{{\mathbf{G}},{{\mathbf{a}}_k}}}). \\[0.6ex]
    \end{gathered}
    \label{endpoints_val}
\end{equation}
For each axis defined by its normal $\mathbf{a}_k$ we compute a 1D GIoU and then combine the three scalar scores through averaging to obtain the final 3D MGIoU:
\begin{equation}
    \begin{gathered}
    {{\mathbf{P}_{{{\mathbf{a}}_k}}}} = [{{\min }_{{\mathbf{P}},{{\mathbf{a}}_k}}},{\max }_{{\mathbf{P}},{{\mathbf{a}}_k}}],{{\mathbf{G}_{{{\mathbf{a}}_k}}}} = [{{\min }_{{\mathbf{G}},{{\mathbf{a}}_k}}},{{{\max }_{{\mathbf{G}},{{\mathbf{a}}_k}}}}], \\
    \left| \mathbf{C} \right| = \max \left( {{{\max }_{{\mathbf{P}},{{\mathbf{a}}_k}}},{{\max }_{{\mathbf{G}},{{\mathbf{a}}_k}}}} \right) - \min \left( {{{\min }_{{\mathbf{P}},{{\mathbf{a}}_k}}},{{\min }_{{\mathbf{G}},{{\mathbf{a}}_k}}}} \right), \\
    \mathrm{GIoU}_k^{1D}=\frac{\left|{{\mathbf{P}_{{{\mathbf{a}}_k}}}}\cap {{\mathbf{G}_{{{\mathbf{a}}_k}}}}\right|}{\left|{{\mathbf{P}_{{{\mathbf{a}}_k}}}}\cup {{\mathbf{G}_{{{\mathbf{a}}_k}}}}\right|}-\frac{\left|\mathbf{C}\right|-\left|{{\mathbf{P}_{{{\mathbf{a}}_k}}}}\cup {{\mathbf{G}_{{{\mathbf{a}}_k}}}}\right|}{\left|\mathbf{C}\right|}, \\
    \mathrm{MGIoU}^{3D}=\frac{1}{3}\sum_{k=1}^3\mathrm{GIoU}_k^{1D}.
    \end{gathered}
    \label{mgiou}
\end{equation}
The final Spatial Point Alignment Loss is:
\begin{equation}
    \mathcal{L}_{3Dcorner}=\frac{1-\mathrm{MGIoU}^{3D}}{2}.
    \label{Spatial Point Alignment Loss}
\end{equation}

\subsection{3D-2D Projection Alignment}
\label{sec:proj alignment}

\begin{figure}[t]
    \centering
    \includegraphics[width=0.95\linewidth]{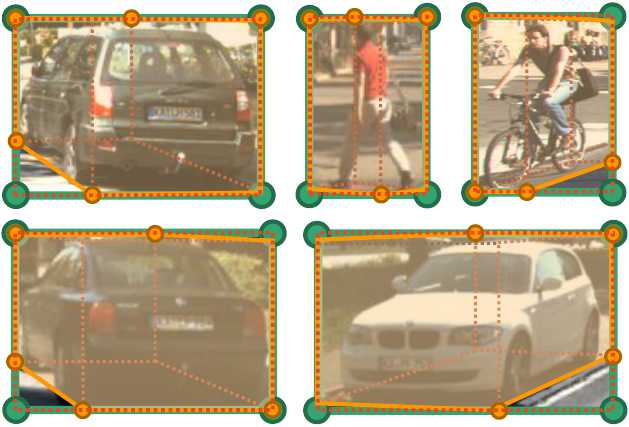}%
    \vspace{-5pt}
    \caption{\textbf{Correspondence between the projected 3D box and the 2D bounding box.} The 2D bounding box is shown as a \textbf{\textcolor[RGB]{52,164,133}{green}} solid rectangle, while the projected region of the 3D box is outlined in \textbf{\textcolor[RGB]{255,152,0}{orange}}. The minimal enclosing rectangle of the projection is indicated by a \textbf{\textcolor[RGB]{233,69,32}{red}} dashed line. As the figure demonstrates, when the eight corner points of the 3D box are projected onto the 2D image plane, at least four of these points are guaranteed to lie on the four boundaries of the 2D bounding box.}
    \label{fig:relation}
    \vspace{-10pt}
 \end{figure}

The perspective projection constraint is a fundamental geometric prior in monocular 3D object detection. The projected silhouette of 3D bounding box should tightly coincide with its corresponding 2D detection box. This constraint was first leveraged by~\cite{deep3dbox} to solve for the 3D center coordinates. In this work, we extend this principle. As illustrated in Fig. \ref{fig:relation}, the projection of a predicted 3D bounding box onto the image plane produces a convex polygon whose horizontally aligned minimal enclosing rectangle ideally coincides exactly with the 2D bounding box. Furthermore, the projected vertices of the 3D box should satisfy the condition that the extreme $u$ and $v$ coordinates lie precisely on the boundaries of the 2D box. A detailed discussion of this constraint is provided in the appendix Sec. \ref{sec:detailed discussion on projection constraint}.

Based on the aforementioned extended perspective projection prior, we propose a 3D-2D Projection Alignment Loss to explicitly constrain geometric consistency. As detailed in Sec. \ref{Spatial Point Alignment}, we first compute the eight corner points $\{\mathbf{P}_i\}_{i=1}^8$ of the predicted 3D box. Then, we apply the camera projection model to obtain their corresponding 2D projections $\{\mathbf{p}_i=(u_i,v_i)\}_{i=1}^8$:
\begin{equation}
    \begin{gathered}
    u_{i}=f_{u}\frac{x_{i}}{z_{i}}+c_{u}, \\
    v_{i}=f_{v}\frac{y_{i}}{z_{i}}+c_{v},
    \end{gathered}
    \label{2D proj points}
\end{equation}
where $f_u$ and $f_v$ are the focal lengths in the $u$ and $v$ directions, respectively, and $c_u$, $c_v$ denote the pixel coordinates of the principal point along the $u$ and $v$ axes.

Next, we compute the minimum and maximum values of the projected $u$ and $v$ coordinates as follows:
\begin{equation}
    \begin{gathered}
    u_{\mathrm{min}}=\min_{1\leq i\leq8}\left(u_{i}\right),v_{\mathrm{min}}=\min_{1\leq i\leq8}\left(v_{i}\right), \\
    u_{\mathrm{max}}=\max_{1\leq i\leq8}\left(u_{i}\right),v_{\mathrm{max}}=\max_{1\leq i\leq8}\left(v_{i}\right).
    \end{gathered}
    \label{projected val}
\end{equation}
We denote the minimal enclosing rectangle of the projected region as $\mathcal{B}_{\mathrm{proj}}^{2D}=[u_{\mathrm{min}},u_{\mathrm{max}}]\times[v_{\mathrm{min}},v_{\mathrm{max}}]$ and the ground-truth 2D bounding box as $\mathcal{B}_{\mathrm{gt}}^{2D}=
\begin{bmatrix}
u_{\mathrm{min}}^{\mathrm{gt}},u_{\mathrm{max}}^{\mathrm{gt}}
\end{bmatrix}\times
\begin{bmatrix}
v_{\mathrm{min}}^{\mathrm{gt}},v_{\mathrm{max}}^{\mathrm{gt}}
\end{bmatrix}$.
The degree of overlap between $\mathcal{B}_{\mathrm{proj}}^{2D}$ and $\mathcal{B}_{\mathrm{gt}}^{2D}$ is then assessed via their 2D GIoU:
\begin{equation}
    \begin{gathered}
    \mathrm{IoU}^{2D}=\frac{\left|\mathcal{B}_{\mathrm{proj}}^{2D}\cap\mathcal{B}_{\mathrm{gt}}^{2D}\right|}{\left|\mathcal{B}_{\mathrm{proj}}^{2D}\cup\mathcal{B}_{\mathrm{gt}}^{2D}\right|}, \\
    \mathrm{GIoU}^{2D}=\mathrm{IoU}^{2D}-\frac{A_{c}-\left|\mathcal{B}_{\mathrm{proj}}^{2D}\cup\mathcal{B}_{\mathrm{gt}}^{2D}\right|}{A_{c}},
    \end{gathered}
    \label{proj giou}
\end{equation}
where $A_c$ denotes the area of the smallest enclosing rectangle covering both $\mathcal{B}_{\mathrm{proj}}^{2D}$ and $\mathcal{B}_{\mathrm{gt}}^{2D}$. The final 3D-2D Projection Alignment Loss is formulated as:
\begin{equation}
    \begin{gathered}
    \mathcal{L}_{proj}=1-\mathrm{GIoU}^{2D}.
    \end{gathered}
\end{equation}

\subsection{Total Loss and Training Strategy}
\label{sec:total loss and htl}

We evaluate the proposed method on three representative monocular 3D detectors \cite{monodetr, movis, monodgp}. The overall loss comprises four components: the 2D regression loss $\mathcal{L}_{2D}$, the 3D regression loss $\mathcal{L}_{3D}$, the object-level depth map loss $\mathcal{L}_{dmap}$, and our proposed geometric collaborative losses $\mathcal{L}_{3Dcorner}$ and $\mathcal{L}_{proj}$. 
Refer to appendix Sec. \ref{sec:detailed loss func} for the detailed design of the loss functions.

Following the original designs of \cite{monodetr}, we inherit their respective weighting schemes for $\mathcal{L}_{2D}$, $\mathcal{L}_{3D}$, and $\mathcal{L}_{dmap}$. Specifically, $\mathcal{L}_{2D}$ includes classification loss, 2D bounding box regression loss, 2D GIoU loss, and projected center loss, while $\mathcal{L}_{3D}$ encompasses dimension loss, orientation loss, and uncertainty-based depth loss. We denote the weights of the individual terms in $\mathcal{L}_{2D}$ and $\mathcal{L}_{3D}$ as $\lambda_1$ to $\lambda_7$, and assign $\lambda_8$ to $\mathcal{L}_{dmap}$. 
Furthermore, in the experiments based on MonoDGP~\cite{monodgp}, we retain the region segmentation loss $\mathcal{L}_{region}$ and reformulate the depth loss following the geometry-error-corrected projection formulation:
\begin{equation}
    \mathcal{L}_{depth}=\frac{\sqrt{2}}{\sigma_{d}}\left\|\frac{f\cdot h_{3d}}{h_{2d}}+Z_{err}-Z_{gt}\right\|_{1}+\log(\sigma_{d}).
    \label{depth loss}
\end{equation}
We subsequently incorporate the Spatial Point Alignment loss $\mathcal{L}_{3Dcorner}$ and the 3D-2D Projection Alignment loss $\mathcal{L}_{proj}$ into the overall training objective, with $\mathcal{L}_{depth}$ already subsumed within $\mathcal{L}_{3D}$, yielding the final loss:
\begin{equation}
    \begin{gathered}
    \mathcal{L}_{total}=\frac{1}{N_{gt}}\sum_{n=1}^{N_{gt}}\left(\mathcal{L}_{2D}+\mathcal{L}_{3D}+\lambda_{c}\mathcal{L}_{3Dcorner}+\lambda_{p}\mathcal{L}_{proj}\right) \\
    +\lambda_{8}\mathcal{L}_{dmap}+\lambda_{9}\mathcal{L}_{region},
    \end{gathered}
    \label{total loss}
\end{equation}
where ${N_{gt}}$ denotes the number of ground-truth objects.

Although SPAN promotes joint optimization of the 3D attributes and markedly improves both spatial and projection consistency, it can destabilize early training.
Specifically, noisy initial predictions of the 3D box parameters can invalidate the assumptions underlying these high‑order constraints. To mitigate this, we adopt and adapt the Hierarchical Task Learning (HTL) strategy~\cite{gupnet}.
HTL applies a time‑varying weight ${\omega _i}\left(t \right)$ to each loss term, dynamically adjusting these weights at each epoch based on the learning progress of their prerequisite tasks.

\begin{figure}[t]
    \centering
    \includegraphics[width=1.0\linewidth]{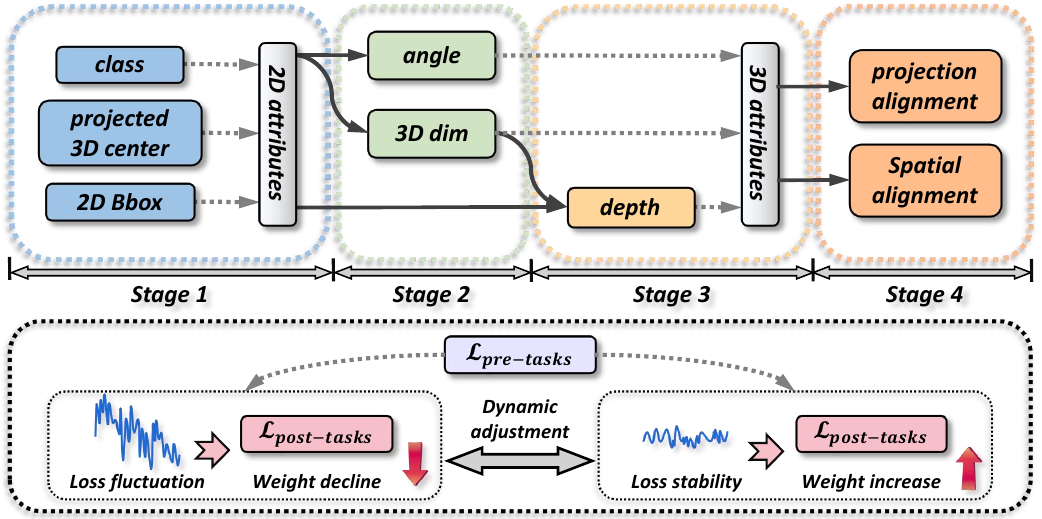}%
    \vspace{-5pt}
    \caption{\textbf{Illustration of the task hierarchy.} The overall training process is divided into four sequential stages. Under the dynamic adjustment of Hierarchical Task Learning, a subsequent stage can only receive a significant loss weight once its pre-tasks have been trained to a stable state.}
    \label{fig:HTL}
    \vspace{-10pt}
 \end{figure}

From a high-level perspective, as illustrated in Fig. \ref{fig:HTL}, the overall learning procedure is organized into four sequential stages.
Stage 1 focuses on 2D detection, covering object classification, 2D box localization, and projected center regression.
Stage 2 addresses 3D dimension and rotation angle regression. Following MonoDGP~\cite{monodgp}, whose decoder is decoupled and initializes 3D object queries from 2D cues, we treat the tasks of Stage 1 as pre-tasks for Stage 2.
Stage 3 addresses depth estimation. As this process relies on the geometric relationship between the 2D box attributes from Stage 1 and the 3D attributes from Stage 2, it is logically dependent on the outputs of both.
Finally, Stage 4 introduces the Spatial-Projection Alignment, which depends on all preceding 3D attribute regression tasks.
This staged design ensures that each task commences only after its prerequisite tasks have stabilized, thereby preserving training stability throughout the optimization process.
The implementation of HTL is provided in the appendix Sec. \ref{Implementation of Hierarchical Task Learning}.
\section{Experiments}

\begin{table*}[!ht]
\centering
\fontsize{8.5pt}{10pt}\selectfont
\begin{tabular}{l|c|c|ccc|ccc|ccc}
\toprule[0.15em]
\rule{0pt}{0.3cm}
\multirow{2}{*}{Methods} & \multirow{2}{*}{Reference} & \multirow{2}{*}{Extra data} & \multicolumn{3}{c|}{Test, $A{P_{3D|R40}}$} & \multicolumn{3}{c|}{Test, $A{P_{BEV|R40}}$} & \multicolumn{3}{c}{Val, $A{P_{3D|R40}}$} \\
& & & Easy & Mod. & Hard & Easy & Mod. & Hard & Easy & Mod. & Hard \\ \hline
CaDDN~\cite{caddn} & CVPR 2021 & & 19.17 & 13.41 & 11.46 & 27.94 & 18.91 & 17.19 & 23.57 & 16.31 & 13.84 \\
CMKD~\cite{hong2022cross} & ECCV 2022 & & 25.09 & 16.99 & 15.30 & 33.69 & 23.10 & 20.67 & - & - & - \\
MonoNeRD~\cite{mononerd} & ICCV 2023 & & 22.75 & 17.13 & 15.63 & 31.13 & 23.46 & 20.97 & 20.64 & 15.44 & 13.99 \\
OccupancyM3D~\cite{occupancym3d} & CVPR 2024 & \multirow{-4}{*}{LiDAR} & 25.55 & 17.02 & 14.79 & \underline{35.38} & 24.18 & 21.37 & 26.87 & 19.96 & 17.15 \\ \hline

MonoPGC~\cite{monopgc} & ICRA 2023 & & 24.68 & 17.17 & 14.14 & 32.50 & 23.14 & 20.30 & 25.67 & 18.63 & 15.65 \\
OPA-3D~\cite{opa-3d} & RAL 2023 & \multirow{-2}{*}{Depth} & 24.60 & 17.05 & 14.25 & 33.54 & 22.53 & 19.22 & 24.97 & 19.40 & 16.59 \\ \hline

GUPNet~\cite{gupnet} & ICCV 2021 & & 20.11 & 14.20 & 11.77 & - & - & - & 22.76 & 16.46 & 13.72 \\
MonoCon~\cite{monocon} & AAAI 2022 & & 22.50 & 16.46 & 13.95 & 31.12 & 22.10 & 19.00 & 26.33 & 19.01 & 15.98 \\
MonoDETR~\cite{monodetr} & ICCV 2023 & & 25.00 & 16.47 & 13.58 & 33.60 & 22.11 & 18.60 & 28.84 & 20.61 & 16.38 \\
MonoCD~\cite{monocd} & CVPR 2024 & & 25.53 & 16.59 & 14.53 & 33.41 & 22.81 & 19.57 & 26.45 & 19.37 & 16.38 \\
FD3D~\cite{fd3d} & AAAI 2024 & & 25.38 & 17.12 & 14.50 & 34.20 & 23.72 & 20.76 & 28.22 & 20.23 & 17.04 \\
GUPNet++~\cite{gupnet++} & TPAMI 2024 & & 24.99 & 16.48 & 14.58 & - & - & - & 29.03 & 20.45 & 17.89 \\
MoVis~\cite{movis} & TIP 2025 & & 23.99 & 17.52 & 14.82 & - & - & - & 28.46 & 20.77 & 17.70 \\
MonoDGP~\cite{monodgp} & CVPR 2025 & \multirow{-8}{*}{None} & \underline{26.35} & \underline{18.72} & \underline{15.97} & 35.24 & \textbf{25.23} & \textbf{22.02} & \underline{30.76} & \underline{22.34} & \underline{19.02} \\ \hline

\rowcolor{cyan!10}
MonoDGP (+SPAN) & - & None & \textbf{27.02} & \textbf{19.30} & \textbf{16.49} & \textbf{35.58} & \underline{24.83} & \underline{21.79} & \textbf{30.98} & \textbf{23.26} & \textbf{20.17} \\
\textit{Improvement} & - & \textit{v.s. second-best} & \textbf{\textcolor[RGB]{65,105,225}{+0.67}} & \textbf{\textcolor[RGB]{65,105,225}{+0.58}} & \textbf{\textcolor[RGB]{65,105,225}{+0.52}} & \textbf{\textcolor[RGB]{65,105,225}{+0.20}} & \textbf{\textcolor[RGB]{220,20,60}{-0.40}} & \textbf{\textcolor[RGB]{220,20,60}{-0.23}} & \textbf{\textcolor[RGB]{65,105,225}{+0.22}} & \textbf{\textcolor[RGB]{65,105,225}{+0.92}} & \textbf{\textcolor[RGB]{65,105,225}{+1.15}}\\
\bottomrule[0.15em]
\end{tabular}
\caption{Comparison with current state-of-the-art methods on Car category on the KITTI val set and test set. The best results are highlighted in \textbf{bold}, the second-best are \underline{underlined}. 
Improvements over the second-best are shown in \textbf{\textcolor[RGB]{65,105,225}{blue}}, while performance degradations relative to the best are indicated in \textbf{\textcolor[RGB]{220,20,60}{red}}.}
\vspace{-10pt}
\label{tab:total res}
\end{table*}

\begin{table}[!ht]
\centering
\setlength{\tabcolsep}{4pt}
\fontsize{7pt}{10pt}\selectfont
\begin{tabular}{l|c|ccc|ccc}
\toprule[0.15em]
\rule{0pt}{0.3cm}
\multirow{3}*{\centering Methods} & \multirow{3}*{\parbox{0.6cm}{\centering Extra data}} & \multicolumn{6}{c}{\centering Test, $A{P_{3D|R40}}$}\\
\cline{3-8}
\rule{0pt}{0.25cm}
& & \multicolumn{3}{c|}{\centering Pedestrian} & \multicolumn{3}{c}{\centering Cyclist} \\
\cline{3-8}
\rule{0pt}{0.25cm}
& & {\centering Easy} & {\centering Mod.} & {\centering Hard} & {\centering Easy} & {\centering Mod.} & {\centering Hard} \\ \hline
CaDDN~\cite{caddn} & & 12.87 & 8.14 & 6.76 & 7.00 & 3.41 & 3.30 \\
OccupancyM3D~\cite{occupancym3d} & \multirow{-2}{*}{LiDAR} & 14.68 & 9.15 & 7.80 & \underline{7.37} & 3.56 & 2.84 \\ \hline

MonoPGC~\cite{monopgc} & Depth & 14.16 & 9.67 & 8.26 & 5.88 & 3.30 & 2.85 \\ \hline

GUPNet~\cite{gupnet} & & 14.72 & 9.53 & 7.87 & 4.18 & 2.65 & 2.09 \\
MonoCon~\cite{monocon} & & 13.10 & 8.41 & 6.94 & 2.80 & 1.92 & 1.55 \\
DEVIANT~\cite{deviant} & & 13.43 & 8.65 & 7.69 & 5.05 & 3.13 & 2.59 \\
MonoDETR~\cite{monodetr} & & 12.65 & 7.19 & 6.72 & 5.12 & 2.74 & 2.02 \\
GUPNet++~\cite{gupnet++} & & 12.45 & 8.13 & 6.91 & 6.71 & \underline{3.91} & \underline{3.80} \\
MonoDGP~\cite{monodgp} & \multirow{-6}{*}{None} & \underline{15.04} & \underline{9.89} & \underline{8.38} & 5.28 & 2.82 & 2.65 \\ \hline

\rowcolor{cyan!10}
MonoDGP (+SPAN) & None & \textbf{16.62} & \textbf{10.54} & \textbf{9.03} & \textbf{8.08} & \textbf{4.78} & \textbf{3.96} \\
\bottomrule[0.15em]
\end{tabular}
\caption{Comparisons of the pedestrian and cyclist categories on the KITTI test set. We \textbf{bold} the best results and \underline{underline} the second-best results.}
\vspace{-8pt}
\label{tab:ped and cyc res}
\end{table}

\begin{table}[!ht]
\centering
\setlength{\tabcolsep}{5pt}
\fontsize{8pt}{11pt}\selectfont
\begin{tabular}{l|ccc|ccc}
\toprule[0.15em]
\rule{0pt}{0.3cm}
\multirow{2}*{Methods} & \multicolumn{3}{c|}{\centering Val, $A{P_{3D|R40}}$} & \multicolumn{3}{c}{Val, $A{P_{BEV|R40}}$} \\
& Easy & Mod. & Hard & Easy & Mod. & Hard \\ \hline

MonoDETR~\cite{monodetr} & 28.84 & 20.61 & 16.38 & 37.86 & 26.95 & 22.80 \\
\rowcolor{cyan!10}
\textbf{+ SPAN} & \textbf{28.99} & \textbf{21.22} & \textbf{17.08} & \textbf{38.24} & \textbf{27.47} & \textbf{23.42} \\
\textit{Improvement} & \textbf{\textcolor[RGB]{65,105,225}{+0.15}} & \textbf{\textcolor[RGB]{65,105,225}{+0.61}} & \textbf{\textcolor[RGB]{65,105,225}{+0.70}} & \textbf{\textcolor[RGB]{65,105,225}{+0.38}} & \textbf{\textcolor[RGB]{65,105,225}{+0.52}} & \textbf{\textcolor[RGB]{65,105,225}{+0.62}} \\
\hline

MoVis~\cite{movis} & 28.46 & 20.77 & 17.70 & 37.52 & 27.00 & 23.34 \\
\rowcolor{cyan!10}
\textbf{+ SPAN} & \textbf{28.65} & \textbf{21.44} & \textbf{18.52} & \textbf{37.95} & \textbf{27.66} & \textbf{24.12} \\ 
\textit{Improvement} & \textbf{\textcolor[RGB]{65,105,225}{+0.19}} & \textbf{\textcolor[RGB]{65,105,225}{+0.67}} & \textbf{\textcolor[RGB]{65,105,225}{+0.82}} & \textbf{\textcolor[RGB]{65,105,225}{+0.43}} & \textbf{\textcolor[RGB]{65,105,225}{+0.66}} & \textbf{\textcolor[RGB]{65,105,225}{+0.78}} \\
\hline

MonoDGP~\cite{monodgp} & 30.76 & 22.34 & 19.02 & 39.40 & 28.20 & 24.42 \\
\rowcolor{cyan!10}
\textbf{+ SPAN} & \textbf{30.98} & \textbf{23.26} & \textbf{20.17} & \textbf{40.06} & \textbf{29.17} & \textbf{25.56} \\
\textit{Improvement} & \textbf{\textcolor[RGB]{65,105,225}{+0.22}} & \textbf{\textcolor[RGB]{65,105,225}{+0.92}} & \textbf{\textcolor[RGB]{65,105,225}{+1.15}} & \textbf{\textcolor[RGB]{65,105,225}{+0.66}} & \textbf{\textcolor[RGB]{65,105,225}{+0.97}} & \textbf{\textcolor[RGB]{65,105,225}{+1.14}} \\

\bottomrule[0.15em]
\end{tabular}
\caption{To comprehensively validate the effectiveness of our approach, we extend the Spatial-Projection Alignment to three distinct baseline models. All metrics are reported on the Car category of the KITTI val set, with performance gains highlighted in \textbf{\textcolor[RGB]{65,105,225}{blue}}.}
\vspace{-12pt}
\label{tab:val for whole baselines}
\end{table}

\subsection{Setup}

\boldparagraph{Dataset.}
We evaluate our method on the widely used KITTI benchmark~\cite{kitti}. The KITTI dataset contains 7,481 training images and 7,518 test images, covering three object categories, Car, Pedestrian, and Cyclist. Each category is further divided into three difficulty levels, Easy, Moderate, and Hard. Following previous work~\cite{chen20153d}, we split the official training set into 3,712 samples for training and 3,769 samples for validation in our experiments.

\boldparagraph{Evaluation Metrics.}
In KITTI, the model's performance is typically measured using average precision in 3D space and bird’s-eye view ($AP_{3D}$ and $AP_{BEV}$) at 40 recall positions following the established protocol~\cite{simonelli2019disentangling}. The KITTI benchmark ranks all methods mainly based on the moderate $AP_{3D}$ metrics of Car category. Each result on the validation set represents the average value obtained from five independent runs using identical settings.

\boldparagraph{Implementation Details.}
As mentioned above, we integrate our method into three distinct baselines \cite{monodetr, movis, monodgp}.
For the KITTI test set, we select MonoDGP~\cite{monodgp}, as it achieves the best performance on the KITTI validation set among our baselines. The weights of losses are set as \{2, 5, 2, 10, 1, 1, 1, 1, 1\} for $\lambda_1$ to $\lambda_{9}$, respectively, while $\lambda_c$ and $\lambda_p$ are both fixed at 1. All experiments are conducted on a single RTX 3090 GPU with a batch size of 8. We utilize the AdamW~\cite{adamW} optimizer with weight decay and set the initial learning rate to 2e-4. 
Note that the three baselines are trained for a different number of epochs.
For MonoDETR~\cite{monodetr}, we train for 215 epochs, reducing the learning rate by a factor of 0.1 at epochs 145 and 185. For MoVis~\cite{movis}, we train for 220 epochs, reducing the learning rate by a factor of 0.1 at epochs 145 and 190. For MonoDGP~\cite{monodgp}, we train for 300 epochs, reducing the learning rate by a factor of 0.5 at epochs 85, 145, 205, and 265.

\subsection{Main Results}
To comprehensively evaluate the performance of the proposed method, we conduct quantitative experiments on the KITTI test and val splits.
More extensive results on the Waymo dataset can be found in Appendix Sec. \ref{sec:Experiments on Waymo Open Dataset}.

\boldparagraph{Results of Car category on the KITTI test and val set.}
As shown in Tab. \ref{tab:total res}, the proposed method achieves the best performance on most metrics for the Car category, without any additional data. 
Compared with the baseline model MonoDGP~\cite{monodgp}, our approach improves the moderate $AP_{3D}$ by 0.58\% on the test set and by 0.92\% on the val set.
Moreover, our method consistently outperforms models trained with extra data. For instance, it surpasses OccupancyM3D~\cite{occupancym3d} by 2.28\% in moderate $AP_{3D}$ on the test set.

\boldparagraph{Pedestrian/Cyclist detection on the KITTI test set.}
We additionally report the detection results for the Pedestrian/Cyclist on the test set in Tab. \ref{tab:ped and cyc res}. 
Specifically, our method outperforms all other models on all evaluation metrics across all difficulty levels.
In particular, our approach surpasses the second-best method by 0.87\% in moderate $AP_{3D}$ for Cyclist. 
These results further demonstrate the generalization capability of the proposed Spatial-Projection Alignment across multiple object categories, effectively improving localization accuracy even for smaller and non-rigid objects such as pedestrians and cyclists.

\boldparagraph{Spatial-Projection Alignment on varied baselines.}
As detailed in Tab. \ref{tab:val for whole baselines}, we extend SPAN to three competitive monocular 3D detectors.
Experimental results on the KITTI validation set demonstrate that SPAN consistently improves performance across different frameworks.
Notably, the performance gains are most significant under the hard level. We attribute this to the fact that challenging instances are more prone to depth ambiguity and localization errors during detection, whereas the geometric consistency enforced by SPAN effectively alleviates these issues.


\begin{figure*}
   \centering
   \subfloat[ MonoDETR]{
        \centering
        \includegraphics[width=0.3283\linewidth]{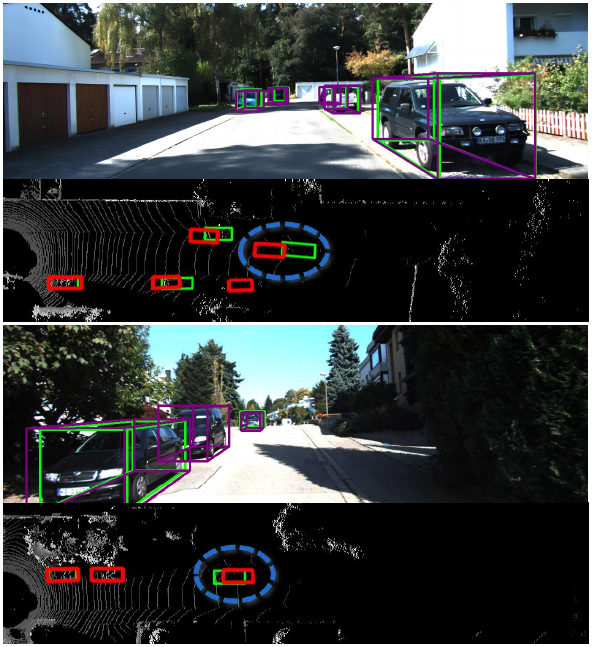}%
    }\hfill
    \subfloat[ MonoDGP]{
        \centering
        \includegraphics[width=0.3283\linewidth]{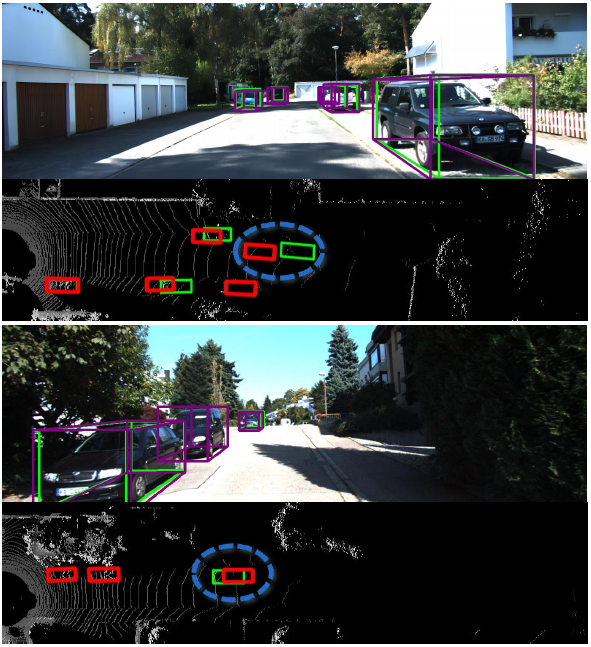}%
    }\hfill
    \subfloat[ MonoDGP (+SPAN)]{
        \centering
        \includegraphics[width=0.3283\linewidth]{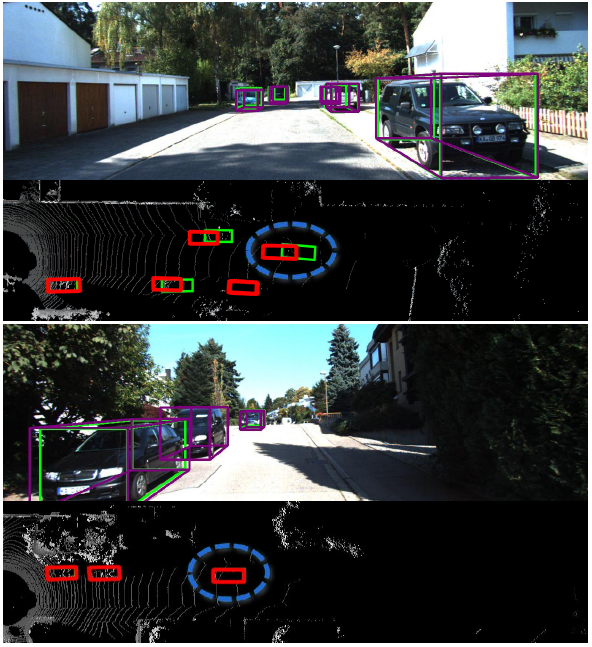}%
    }\hfill
   \vspace{-6pt}
   \caption{\textbf{Qualitative results on KITTI val set.} (a) MonoDETR~\cite{monodetr}, (b) MonoDGP~\cite{monodgp}, and (c) MonoDGP (+SPAN). For each image set, the top row presents the camera-view visualization, while the bottom row offers the corresponding bird’s-eye view. Ground-truth bounding boxes are rendered in \textbf{\textcolor[RGB]{52,164,133}{green}}, and predictions are shown in \textbf{\textcolor[RGB]{233,69,32}{red}}. We also circle some objects to highlight the difference between the baseline model and our method.}
   \vspace{-8pt}
   \label{fig:qualitative res}
\end{figure*}

\begin{table}[!ht]
\centering
\setlength{\tabcolsep}{8pt}
\fontsize{8.5pt}{8.5pt}\selectfont
\begin{tabular}{ccc|ccc}
\toprule[0.15em]
\rule{0pt}{0.3cm}
\multirow{2}{*}{$\mathcal{L}_{3Dcorner}$} & \multirow{2}{*}{$\mathcal{L}_{proj}$} & \multirow{2}{*}{HTL} & \multicolumn{3}{c}{Val, $AP_{3D|R40}$} \\ 
\cline{4-6} 
&  &  & Easy & Mod. & Hard \\ 
\midrule
\XSolidBrush & \XSolidBrush & \XSolidBrush & 30.76 & 22.34 & 19.02 \\
\Checkmark & \XSolidBrush & \XSolidBrush & 29.73 & 21.92 & 18.82 \\
\XSolidBrush & \Checkmark & \XSolidBrush & 29.03 & 21.80 & 18.97 \\
\XSolidBrush & \XSolidBrush & \Checkmark & 30.07 & 22.56 & 19.36 \\
\Checkmark & \XSolidBrush & \Checkmark & \textbf{31.12} & 22.89 & 19.77 \\
\XSolidBrush & \Checkmark & \Checkmark & 30.69 & 22.97 & 19.72 \\
\rowcolor{cyan!10}
\Checkmark & \Checkmark & \Checkmark & 30.98 & \textbf{23.26} & \textbf{20.17} \\ \hline

\multicolumn{3}{c|}{\textit{Improvement}} & \textbf{\textcolor[RGB]{65,105,225}{+0.36}} & \textbf{\textcolor[RGB]{65,105,225}{+0.92}} & \textbf{\textcolor[RGB]{65,105,225}{+1.15}} \\

\bottomrule[0.15em]
\end{tabular}
\caption{Ablation study on different components of our proposed method on the KITTI val set. $\mathcal{L}_{3Dcorner}$ and $\mathcal{L}_{proj}$ denote Spatial Point Alignment loss and 3D-2D Projection Alignment loss, respectively. HTL denotes Hierarchical Task Learning strategy.}
\vspace{-10pt}
\label{tab:overall ablation}
\end{table}

\subsection{Ablation Study}
In our ablation study, MonoDGP~\cite{monodgp} serves as the baseline. The default evaluation metric is the $AP_{3D}$ of the Car category on the KITTI validation set.

\boldparagraph{Overall performance.}
Tab. \ref{tab:overall ablation} summarizes the ablative analysis, where we systematically evaluate the contribution of each component. Our proposed method introduces three key techniques: Spatial Point Alignment, 3D-2D Projection Alignment, and the Hierarchical Task Learning strategy. Compared to the baseline, the best $AP_{3D}$ under
three-level difficulties gain an increase of 0.36\%, 0.92\% and 1.15\%, respectively. These experiments validate the
effectiveness of each technique utilized in our proposed method.

\begin{table}[!t]
\centering
\setlength{\tabcolsep}{15pt}
\fontsize{8.5pt}{10pt}\selectfont
\begin{tabular}{cc|ccc}
\toprule[0.15em]
\multicolumn{2}{c|}{Loss weight} & \multicolumn{3}{c}{Val, $AP_{3D|R40}$} \\ \hline
$\lambda_c$ & $\lambda_p$ & Easy & Mod. & Hard \\
\midrule
0.5 & 0.5 & 30.50 & 22.81 & 19.60 \\
0.5 & 1.0 & 30.75 & 22.98 & 19.89 \\
1.0 & 0.5 & 30.85 & 23.01 & 19.86 \\
\rowcolor{cyan!10}
1.0 & 1.0 & \textbf{30.98} & \textbf{23.26} & \textbf{20.17} \\
1.0 & 2.0 & 30.62 & 22.75 & 19.44 \\
2.0 & 1.0 & 30.53 & 22.86 & 19.51 \\
2.0 & 2.0 & 30.37 & 22.66 & 19.34 \\

\bottomrule[0.15em]
\end{tabular}
\caption{Ablation study on the loss weights of the Spatial-Projection Alignment on the KITTI val set. $\lambda_c$ and $\lambda_p$ denote the weights of the Spatial Point Alignment loss and the 3D-2D Projection Alignment loss, respectively.}
\vspace{-10pt}
\label{tab:loss weights}
\end{table}

It is worth noting that applying the Spatial Point Alignment constraint or the 3D-2D Projection Alignment constraint alone without the support of Hierarchical Task Learning strategy actually reduces model performance. 
This finding supports our discussion in Sec. \ref{sec:total loss and htl}, where we show that directly enforcing geometric collaborative constraints during the early stages of training leads to instability and ultimately harms overall effectiveness. 
Only by integrating these constraints within the Hierarchical Task Learning framework can their full potential be realized.

\boldparagraph{Loss weights.}
We further evaluated how varying the weights of the Spatial Point Alignment loss and the 3D-2D Projection Alignment loss affects overall performance. Their corresponding loss weights are denoted as $\lambda_c$ and $\lambda_p$, respectively.
Tab. \ref{tab:loss weights} summarizes the performance trend of our proposed losses under various weightings.
The best overall result is achieved when $\lambda_c$ and $\lambda_p$ are both set to 1.0.
Increasing these weights from 0.5 to 1.0 produces an approximately 0.5\% percent gain in $AP_{3D}$ across all three difficulty levels. 
This improvement can be attributed to the strengthened geometric consistency, which enforces tighter alignment in both 3D space and 2D projection, thereby enhancing localization accuracy.
However, further amplifying $\lambda_c$ and $\lambda_p$ beyond 1.0 degrades performance, presumably because the inflated loss terms begin to dominate the total objective and suppress the influence of the core 3D regression losses during later training stages, leading to accumulated prediction errors.

\subsection{Robustness to 2D Detection Errors}
\label{sec:robustness}
We conducted a robustness test for the proposed scheme by simulating different levels of 2D detection noise. 
Specifically, during the training phase, we added uniform random noise to the ground truth coordinates of each corner of the 2D bounding box to simulate real-world scenarios such as camera shake, sensor noise, or imperfect 2D detector outputs. Tab. \ref{tab:2d noise} shows the performance degradation of SPAN at five different noise levels.

The experimental data shows that SPAN maintains a reasonable performance when the perturbation is less than 10px, which demonstrates that its geometric constraint mechanism makes it robust to 2D detection noise to a certain extent. 
This resilience stems from the fact that the geometric constraints enforce collective consistency between all corner points of the bounding box. Small errors in individual corner detections can be effectively averaged out or compensated for by the constraints imposed by the overall geometric structure, thereby preserving a stable 3D-to-2D mapping. 
It is worth noting that when the perturbation is greater than 15px, the performance of SPAN drops sharply. 
We believe that such a large 2D localization error has exceeded the compensation capability of the geometric constraints, fundamentally undermining the assumption of 3D-2D correspondence.

\begin{table}[!ht]
\centering
\setlength{\tabcolsep}{5pt}
\fontsize{8.5pt}{12pt}\selectfont
\begin{tabular}{l|ccc|ccc}
\toprule[0.15em]
\rule{0pt}{0.3cm}
\multirow{2}{*}{Noise Level} & \multicolumn{3}{c|}{$AP_{3D|R40}$} & \multicolumn{3}{c}{$AP_{BEV|R40}$} \\
\cline{2-7}
& Easy & Mod. & Hard & Easy & Mod. & Hard \\ \hline
\rowcolor{cyan!10}

Clean & \textbf{30.98} & \textbf{23.26} & \textbf{20.17} & 40.06 & \textbf{29.17} & \textbf{25.56} \\
$\pm$2px & 30.91 & 22.65 & 19.92 & \textbf{40.18} & 28.74 & 24.93 \\
$\pm$5px & 28.97 & 21.23 & 18.44 & 37.75 & 27.19 & 23.42 \\
$\pm$10px & 26.38 & 18.89 & 16.47 & 34.67 & 24.45 & 21.18 \\
\rowcolor{red!10}
$\pm$15px & 19.76 & 13.51 & 11.73 & 26.28 & 17.93 & 15.68 \\
\bottomrule[0.15em]
\end{tabular}
\caption{Robustness analysis of SPAN against 2D detection noise on KITTI val set. We introduce uniform random noise to each 2D bounding box corner coordinate during training.}
\vspace{-6pt}
\label{tab:2d noise}
\end{table}



\subsection{Qualitative Results}
To facilitate an intuitive comparison between our approach and the baseline, we provide qualitative visualizations in Fig. \ref{fig:qualitative res}.
Equipping the baseline MonoDGP~\cite{monodgp} with the proposed Spatial-Projection Alignment markedly improves localization accuracy for most objects in the scene.
This improvement is particularly evident on challenging samples, such as the distant vehicles, where the geometric collaborative constraints help the monocular 3D detector achieve more accurate localization.
\section{Conclusion}
In this paper, we re-examine a fundamental limitation of existing monocular 3D detectors, the neglect of the intrinsic geometric and projective relationships among bounding-box attributes.
To address this oversight, we introduce Spatial-Projection Alignment (SPAN), which jointly enforce 3D corner alignment and 3D-2D projection alignment, thereby guaranteeing spatial coherence of every predicted 3D box and its consistency with the corresponding 2D detection.
These constraints are then integrated into a tailored Hierarchical Task Learning strategy, ensuring stable optimization throughout training.
~Extensive experiments demonstrate that SPAN can be integrated seamlessly into existing pipelines without any architectural changes or added inference cost while still improving baseline performance.
This study underscores the value of explicit geometric regularization in monocular 3D perception and we hope to extend this approach to multi‑view 3D perception in future work.
{
    \small
    \bibliographystyle{ieeenat_fullname}
    \bibliography{main}
}

\clearpage
\setcounter{page}{1}
\maketitlesupplementary
\setcounter{section}{0}
\renewcommand{\thesection}{\Alph{section}}

\begin{strip}
    \vspace{-20pt}
    \subsection*{Content}
    
    This Appendix contains the following parts:
    \begin{itemize}
        \item \textbf{Detailed Discussion on Projection Constraint}. We present a detailed discussion from both geometric and statistical perspectives, elucidating why the projection constraint holds.
        \item \textbf{Detailed Loss Function}. We delineate the training loss functions in detail, including all components of the 2D and 3D regression losses.
        \item \textbf{Implementation of Hierarchical Task Learning}. We provide a detailed implementation of our Hierarchical Task Learning strategy.
        \item \textbf{Experiments on Waymo Open Dataset}. We demonstrate the superior performance of our proposed method on Waymo Open Dataset.
        \item \textbf{Compared with Prior Geometric Methods}. We compare SPAN with prior geometric approaches, highlighting the advantages of our differentiable soft constraints over hard algebraic solving.
        \item \textbf{Why MGIoU?} We provide ablation study to justify the choice of MGIoU and demonstrate its optimization dynamics.
        \item \textbf{Disentangling HTL \& Geometric Alignment Gains}. We conduct experiments to isolate the contributions of HTL and geometric alignment losses.
        \item \textbf{Interplay between Projection Alignment \& Depth Bias}. We analyze the interaction between projection alignment and depth bias, showing how they mutually regularize each other.
        \item \textbf{More Qualitative Results}. We present more qualitative results, illustrating that our proposed method achieves better localization for distant objects.
    \end{itemize}
    \bigskip

    \hrulefill
    \vspace*{4pt} 
\end{strip}

\section{Detailed Discussion on Projection Constraint}
\label{sec:detailed discussion on projection constraint}
In Sec. \ref{sec:proj alignment} we extend the projection constraint originally proposed in Deep3DBox~\cite{deep3dbox}, which stipulates that the projected vertices of a 3D bounding box must satisfy the condition that their extreme $u$ and $v$ coordinates coincide exactly with the boundaries of the associated 2D box. We now provide a formal justification for this requirement.

Let $S$ denote the set of points enclosed by a 3D bounding box, and let $\mathbf{P}_n \in S$ be an arbitrary point therein. Under the assumption that the camera intrinsics remain constant, the projection of $\mathbf{P}_n$ onto the 2D image plane is given by $\mathbf{p}_n=(u_n,v_n)$.
\begin{equation}
    \begin{gathered}
    {u_n} = {f_u}\frac{{{x_n}}}{{{z_n}}} + {c_u} = {f_u}f\left( {{x_n},{z_n}} \right) + {c_u}, \hfill \\
    {v_n} = {f_v}\frac{{{y_n}}}{{{z_n}}} + {c_v} = {f_v}f\left( {{y_n},{z_n}} \right) + {c_v}, \hfill \\ 
    \end{gathered}
\end{equation}
with $(x_n,y_n,z_n)$ being the coordinates of $\mathbf{P}_n$ in the camera frame.
We first aim to demonstrate that the extrema of $u_n$ over the point set $S$ can be attained at the vertices of the 3D bounding box, and the same holds for $v_n$.

\subsection{Problem simplification.}
\label{Problem simplification}
The original claim reduces to demonstrating that the extremum of the function $f\left( {{x_n},{z_n}} \right)$ over the rectangular domain $\mathcal{D}_{xz}\{(x,z)|x_{\min}\leq x_{n}\leq x_{\max},0<z_{\min}\leq z_{n}\leq z_{\max}\}$ is attained at one of its four corner points.

\subsection{No interior extrema.}
Within the domain $\mathcal{D}_{xz}$, the gradient of $f\left( {{x_n},{z_n}} \right)$ is expressed as:
\begin{equation}
    \nabla f=\left(\frac{\partial f}{\partial x_n},\frac{\partial f}{\partial z_n}\right)=\left(\frac{1}{z_n},-\frac{x_n}{z_n^2}\right),
\end{equation}
since the gradient possesses neither zeros nor singularities anywhere in $\mathcal{D}_{xz}$, $f\left( {{x_n},{z_n}} \right)$ admits no interior critical points, hence its extrema must occur on the boundary of the domain.

\subsection{Boundary extremum analysis.}
When $z_n$ is fixed at $z_{\min}$ or $z_{\max}$, $f$ reduces to a linear-fractional function of $x_n$ whose maximum and minimum are attained at $x_{\min}$ and $x_{\max}$.
When $x_n$ is fixed at $x_{\min}$ or $x_{\max}$, $f$ reduces to a linear-fractional function of $z_n$ whose maximum and minimum are attained at $z_{\min}$ and $z_{\max}$.

In summary, the extrema of $f\left( {{x_n},{z_n}} \right)$ over the point set $S$ are necessarily attained in at least one of the four line segments $(x_{\min}, y_n, z_{\min})$, $(x_{\min}, y_n, z_{\max})$, $(x_{\max}, y_n, z_{\min})$, and $(x_{\max}, y_n, z_{\max})$.
Since the vertices of the 3D bounding box are precisely the endpoints of these segments, the maximum and minimum values of $u_n$ over $S$ are realized at these vertices, the identical argument holds for $v_n$.

\subsection{Perspective projection preserves convexity.}
Perspective projection is a convexity-preserving linear mapping in homogeneous coordinates. Consequently, the image of any convex set under projection remains convex:
\begin{equation}
    \pi\left(\mathrm{conv}(S)\right)=\mathrm{conv}\left(\pi(S)\right),
\end{equation}
where $\pi$ denotes the projection transformation. The 2D point set obtained by projecting $S$ forms a convex polygon, and every point within this polygon can be represented as a convex combination of its vertices. Therefore:
\begin{equation}
    {u_{\min }} \leqslant {u_n} \leqslant {u_{\max }},{v_{\min }} \leqslant {v_n} \leqslant {v_{\max }}.
\end{equation}

We denote the ground-truth 2D bounding box as $\mathcal{B}^{2D}=
\begin{bmatrix}
u_{\min}^{2D},u_{\max}^{2D}
\end{bmatrix}\times
\begin{bmatrix}
v_{\min}^{2D},v_{\max}^{2D}
\end{bmatrix}$. Since $\mathcal{B}^{2D}$ is the axis‐aligned minimal enclosing rectangle of the projected 3D box point set, it follows that:
\begin{equation}
    \begin{gathered}
    u_{\mathrm{min}}^{2D}=\mathrm{min}u_{n}=u_{\mathrm{min}},u_{\mathrm{max}}^{2D}=\mathrm{max}u_{n}=u_{\mathrm{max}}, \\
    v_{\mathrm{min}}^{2D}=\min v_{n}=v_{\mathrm{min}},v_{\mathrm{max}}^{2D}=\max v_{n}=v_{\mathrm{max}}.
    \end{gathered}
\end{equation}
Consequently, the extrema of the projected point set’s $u$ and $v$ coordinates lie exactly at the edges of the corresponding 2D bounding box. Since we have already shown that these extrema occur at the vertices of the 3D box, it follows that the projected corners of the 3D box provide the extreme $u$ and $v$ values aligned with the boundary of the 2D box, which completes the proof.

\subsection{Statistical analysis.}

\begin{figure*}[t]
    \centering
    \includegraphics[width=0.8\linewidth]{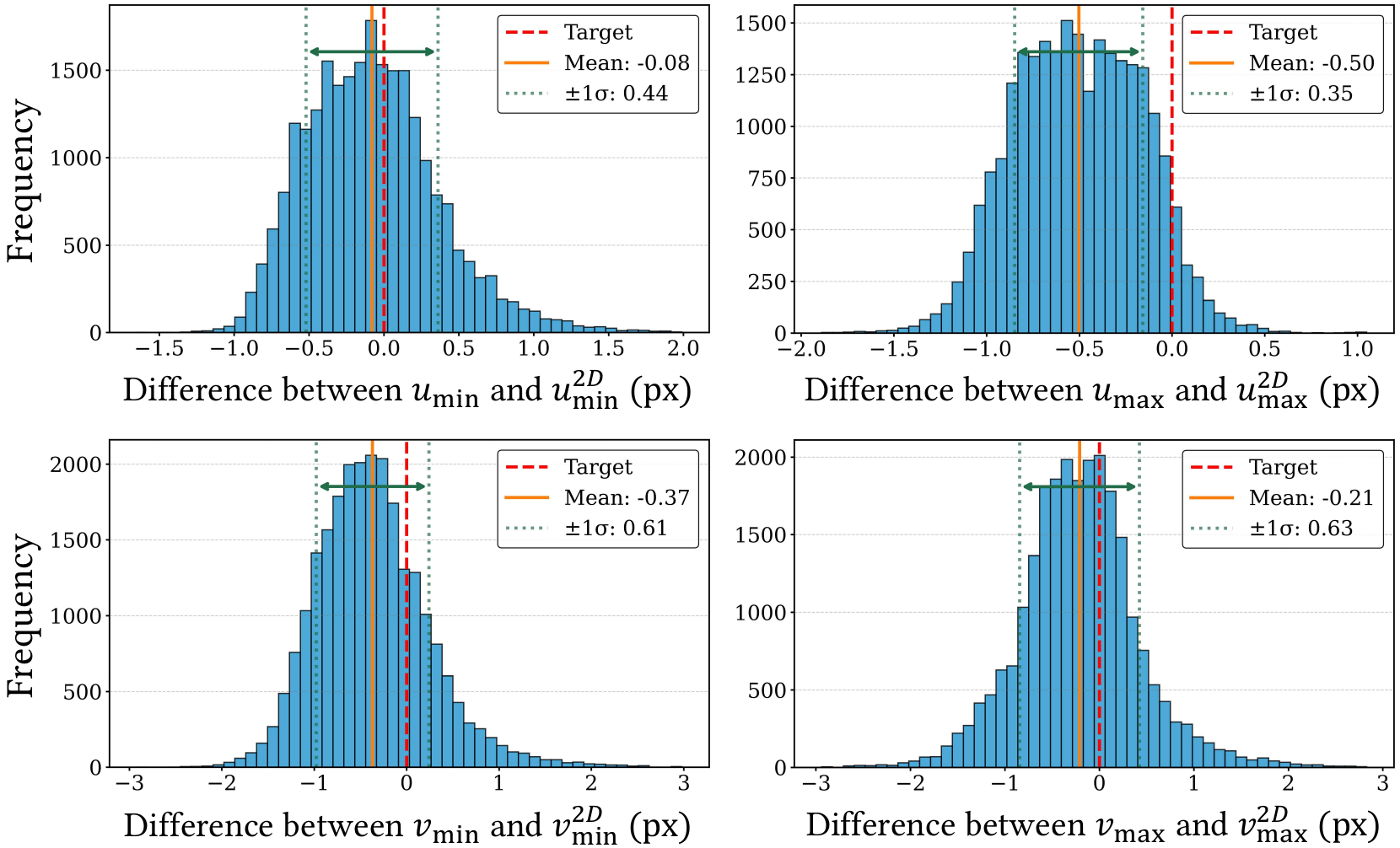}%
    \caption{\textbf{Deviation statistics.} The data distribution of pixel deviation between the extreme projected coordinates of 3D bounding box corners and the corresponding 2D boundaries on the KITTI training set.
    Four histograms report the differences between (i) $u_{\mathrm{min}}$ and $u_{\mathrm{min}}^{2D}$, (ii) $u_{\mathrm{max}}$ and $u_{\mathrm{max}}^{2D}$, (iii) $v_{\mathrm{min}}$ and $v_{\mathrm{min}}^{2D}$, and (iv) $v_{\mathrm{max}}$ and $v_{\mathrm{max}}^{2D}$, where $u_{\mathrm{min}}$, $u_{\mathrm{max}}$ and $v_{\mathrm{min}}$, $v_{\mathrm{max}}$ denote the minimal and maximal projected $u$, $v$ values among the eight 3D box vertices, and $\begin{bmatrix}u_{\min}^{2D},u_{\max}^{2D}\end{bmatrix}\times\begin{bmatrix}v_{\min}^{2D},v_{\max}^{2D}\end{bmatrix}$ denotes the corresponding 2D bounding box.}
    \label{fig:distribution}
    \vspace{-6pt}
 \end{figure*}

Fig.~\ref{fig:distribution} illustrates the distribution of pixel deviations between the edges of the minimal enclosing rectangle of the projected ground-truth 3D bounding box corners and the corresponding edges of the 2D ground‐truth detection boxes on the KITTI training set.
Specifically, we compute the absolute errors $\Delta u_{\min}$, $\Delta u_{\max}$, $\Delta v_{\min}$, and $\Delta v_{\max}$ along the four sides and display their empirical frequency distributions.

All four histograms exhibit a sharp unimodal shape, with the absolute value of each mean below one pixel. Moreover, the worst-case deviation between $v_{\mathrm{min}}$ and $v_{\mathrm{min}}^{2D}$, as well as between $v_{\mathrm{max}}$ and $v_{\mathrm{max}}^{2D}$, is approximately three pixels. 
These statistical observations corroborate the validity of our adopted projection consistency constraint, whereby the minimum bounding rectangle of the projected 3D box aligns almost perfectly with its 2D counterpart.

\section{Detailed Loss Function}
\label{sec:detailed loss func}
For the 2D regression loss $\mathcal{L}_{2D}$, we adopt the focal loss~\cite{focal_loss} to supervise the learning of object categories, while an L1 loss is employed to constrain the projected 3D center $(u_{3d},v_{3d})$ and the dimensions of the 2D bounding box. To impose additional geometric regularization on the 2D box localization, we further incorporate the GIoU loss. Consequently, the 2D regression loss can be formulated as:
\begin{equation}
    \begin{gathered}
    \mathcal{L}_{2D}=\lambda_{1}\mathcal{L}_{class}+\lambda_{2}\mathcal{L}_{2Ddim}+\lambda_{3}\mathcal{L}_{center}+\lambda_{4}\mathcal{L}_{giou}.
    \end{gathered}
    \label{2D loss}
\end{equation}

The 3D regression loss follows the design philosophy of MonoDLE~\cite{monodle}, which uses a normalized L1 loss to supervise the recovery of 3D dimensions $(h_{3d},w_{3d},l_{3d})$, while the yaw angler $r_y$ is regularized by a multi-bin classification loss \cite{deep3dbox}.
For depth estimation, we adopt the uncertainty-aware Laplace-based regression loss proposed in MonoDGP~\cite{monodgp}:
\begin{equation}
    \mathcal{L}_{depth}=\frac{\sqrt{2}}{\sigma_{d}}\left\|\frac{f\cdot h_{3d}}{h_{2d}}+Z_{err}-Z_{gt}\right\|_{1}+\log(\sigma_{d}),
    \label{depth loss2}
\end{equation}
where $\sigma_d$ denotes the standard deviation of the Laplace distribution. Consequently, the 3D regression loss can be compactly expressed as:
\begin{equation}
    \begin{gathered}
    \mathcal{L}_{3D}=\lambda_{5}\mathcal{L}_{3Ddim}+\lambda_{6}\mathcal{L}_{angle}+\lambda_{7}\mathcal{L}_{depth}.
    \end{gathered}
    \label{3D loss}
\end{equation}

The object-level depth map loss, denoted as $\mathcal{L}_{dmap}$, employs the focal loss~\cite{focal_loss} to supervise the classification of foreground depth pixels. The construction of the corresponding ground-truth foreground depth map follows the protocol detailed in MonoDETR~\cite{monodetr}. For the region segmentation loss $\mathcal{L}_{region}$, we adopt the Dice loss introduced in MonoDGP~\cite{monodgp}, which effectively alleviates class-imbalance issues during training:
\begin{equation}
    \begin{gathered}
    \mathcal{L}_{region}^i=1-\frac{2\cdot\sum_{j=1}^{N_i}p_j\cdot g_j}{\sum_{j=1}^{N_i}p_j+\sum_{j=1}^{N_i}g_j}, \\
    \mathcal{L}_{region} = \sum_{i=1}^{r}{\mathcal{L}_{region}^i},
    \end{gathered}
    \label{dice loss}
\end{equation}
where $p_j$ denotes the predicted probability of the $j$-th pixel, $g_j$ is its corresponding binary label in the ground truth mask, and $N_i$ indicates the total number of pixels in the $i$-th feature map $f_i$, $i=1,2,3,4$.

\section{Implementation of Hierarchical Task Learning}
\label{Implementation of Hierarchical Task Learning}
As outlined in Sec. \ref{sec:total loss and htl}, we decompose the training procedure of a monocular 3D object detector into a sequence of pre- and post-dependent tasks. We now elaborate on the refined implementation of the Hierarchical Task Learning (HTL) mechanism, which dynamically modulates the per-epoch weight of each task. Under this scheme, the overall loss can be expressed as:
\begin{equation}
    \mathcal{L}=\sum_{i\in\mathcal{T}}\lambda_i\omega_i(t)\mathcal{L}_i,
\end{equation}
where $\mathcal{T}$ denotes the set of all tasks and $i\in\mathcal{T}$ index of an individual task. Let $t$ be the current training epoch. $\mathcal{L}_i$ represents the loss function of the $i$-th task, while $\lambda_i$ is its fixed base weight and $\omega_i(t)$ is the adaptive scaling factor assigned during training.

We adopt a polynomial time-scheduling function to govern the evolution of the dynamic weights:
\begin{equation}
    \omega_i(t)=\left(\frac{t}{T_e}\right)^{1-\alpha_i(t)},\alpha_i(t)\in[0,1],
\end{equation}
where $T_e$ denotes the total number of training epochs, and the normalized time variable $\frac{t}{T_e}$ automatically rescales the temporal axis. $\alpha_i(t)$ is the adjustment coefficient at epoch $t$, corresponding to each pre-task for the $i$‑th task.

We now turn to the specification of the scheduling parameter $\alpha_i$:
\begin{equation}
    \alpha_i(t)=\left(\prod_{j\in\mathrm{P}_i}l_{s_j}(t)\right)^{\frac{1}{|\mathrm{P}_i|}},|\mathrm{P}_i|\geq1,
    \label{alpha i}
\end{equation}
where $\mathrm{P}_i$ denote the set of pre-tasks for task $i$, and let $l_{s_j}\in[0,1]$ be the learning-status indicator for task $j \in \mathrm{P}_i$. Eqn.~\ref{alpha i} implies that $\alpha_i$ attains a high value only when every pre-task has been sufficiently mastered.
In contrast to the arithmetic-mean implementation adopted in~\cite{gupnet}, we employ the geometric mean.
The formulation naturally satisfies the boundary conditions (all $l_{{s_j}} = 1 \Rightarrow {\alpha _i} = 1$, all ${l_{{s_j}}} = 0 \Rightarrow {\alpha _i} = 0$) and imposes a sharper suppression when any single $l_{s_j}$ deviates from unity. 
The corresponding gradient flow can be derived as:
\begin{equation}
    \begin{gathered}
    \frac{\partial\alpha_i}{\partial l_{s_k}}=\frac{\alpha_i}{|\mathrm{P}_i|l_{s_k}}=\frac{\left(\prod_{j\in\mathrm{P}_i}l_{s_j}(t)\right)^{\frac{1}{|\mathrm{P}_i|}}}{|\mathrm{P}_i|l_{s_k}}=\\
    \frac{\left(\prod_{j\in\mathrm{P}_i,j\neq k}l_{s_j}(t)\right)^{\frac{1}{|\mathrm{P}_i|}}}{|\mathrm{P}_i|\left(l_{s_k}\right)^{1-\frac{1}{|\mathrm{P}_i|}}},
    \end{gathered}
    \label{grad alpha i}
\end{equation}
when a particular prerequisite indicator $l_{s_j}$ is small, the gradient contribution of that term in Eqn. \ref{grad alpha i} is amplified, thereby providing stronger corrective feedback.
This design balances the influence of all prerequisites while mitigating the ``single failure to zero" pathology inherent in pure product scheduling.

The learning-status indicator $l_{s_{j}}(t)$ is computed as:
\begin{equation}
    \begin{gathered}
    l_{s_{j}}(t)=\\
    \mathrm{clamp}\left(\frac{\mathcal{DF}_{j}(K)-\mathcal{DF}_{j}(t)}{\mathcal{DF}_{j}(K)}, \mathrm{min}=0, \mathrm{max}=1\right), \\
    \mathcal{DF}_{j}(t)=\frac{1}{K}\sum_{\hat{t}=t-K}^{t-1}\left|\dot{\mathcal{L}}_{j}(\hat{t})\right|,
    \end{gathered}
    \label{lsjt}
\end{equation}
in Eqn. \ref{lsjt}, $\dot{\mathcal{L}}_{j}(\hat{t})$ denotes the derivative of the loss function $\mathcal{L}_{j}\left(  \cdot  \right)$ at epoch $\hat{t}$, capturing the instantaneous local trend of task $j$. $\mathcal{DF}_{j}(t)$ is defined as the moving average of these derivatives over the most recent $K$ epochs preceding epoch $t$, thereby summarizing the short-term convergence behavior. When the loss of task $j$ stabilizes, $l_{s_{j}}$ approaches one, indicating that the prerequisite is reliably learned.

\section{Experiments on Waymo Open Dataset}
\label{sec:Experiments on Waymo Open Dataset}

\begin{table*}[!ht]
\fontsize{8pt}{8pt}\selectfont
\centering
\tabcolsep=0.02\linewidth 
\begin{tabular}{c|l|c|cccc}
\toprule[0.15em]
\rule{0pt}{0.35cm}
\multirow{2}*{Difficulty} & \multirow{2}*{Method} & \multirow{2}*{\makecell{Extra \\ Data}} & \multicolumn{4}{c}{$AP_{3D}$} \\ \cline{4-7}
\rule{0pt}{0.35cm}
 & & & All & 0-30 & 30-50 & 50-$\infty$ \\
\hline
\rule{0pt}{0.3cm}
\multirow{7}*{\makecell{Level\_1 \\ ($\mathrm{IoU}_{3D}\ge0.7$)}} & CaDDN \cite{caddn} & LIDAR & 5.03 & 15.54 & 1.47 & 0.10 \\
\rule{0pt}{0.3cm}
 & PCT \cite{pct} & Depth & 0.89 & 3.18 & 0.27 & 0.07 \\
 \rule{0pt}{0.3cm}
 & GUPNet \cite{gupnet} in \cite{deviant} & None & 2.28 & 6.15 & 0.81 & 0.03 \\
 \rule{0pt}{0.3cm}
 & DEVIANT \cite{deviant} & None & 2.69 & 6.95 & 0.99 & 0.02 \\
 \rule{0pt}{0.3cm}
 & MonoUNI \cite{monouni} & None & 3.20 & 8.61 & 0.87 & 0.13 \\
 \rule{0pt}{0.3cm}
 & MonoDGP \cite{monodgp} & None & \underline{4.28} & \underline{10.24} & \underline{1.15} & \underline{0.16} \\
 \rule{0pt}{0.3cm}
 & \textbf{MonoDGP + SPAN} & None & \textbf{4.36} & \textbf{10.43} & \textbf{1.17} & \textbf{0.17} \\
\hline
\rule{0pt}{0.3cm}
\multirow{7}*{\makecell{Level\_2 \\ ($\mathrm{IoU}_{3D}\ge0.7$)}} & CaDDN \cite{caddn} & LIDAR & 4.49 & 14.50 & 1.42 & 0.09 \\
\rule{0pt}{0.3cm}
 & PCT \cite{pct} & Depth & 0.66 & 3.18 & 0.27 & 0.07 \\
 \rule{0pt}{0.3cm}
 & GUPNet \cite{gupnet} in \cite{deviant} & None & 2.14 & 6.13 & 0.78 & 0.02 \\
 \rule{0pt}{0.3cm}
 & DEVIANT \cite{deviant} & None & 2.52 & 6.93 & 0.95 & 0.02 \\
 \rule{0pt}{0.3cm}
 & MonoUNI \cite{monouni} & None & 3.04 & 8.59 & 0.85 & 0.12 \\
 \rule{0pt}{0.3cm}
 & MonoDGP \cite{monodgp} & None & \underline{4.00} & \underline{10.20} & \underline{1.13} & \underline{0.15} \\
 \rule{0pt}{0.3cm}
 & \textbf{MonoDGP + SPAN} & None & \textbf{4.12} & \textbf{10.51} & \textbf{1.16} & \textbf{0.16} \\
\hline
\rule{0pt}{0.3cm}
\multirow{7}*{\makecell{Level\_1 \\ ($\mathrm{IoU}_{3D}\ge0.5$)}} & CaDDN \cite{caddn} & LIDAR & 17.54 & 45.00 & 9.24 & 0.64 \\
\rule{0pt}{0.3cm}
 & PCT \cite{pct} & Depth & 4.20 & 14.70 & 1.78 & 0.39 \\
 \rule{0pt}{0.3cm}
 & GUPNet \cite{gupnet} in \cite{deviant} & None & 10.02 & 24.78 & 4.84 & 0.22 \\
 \rule{0pt}{0.3cm}
 & DEVIANT \cite{deviant} & None & 10.98 & 26.85 & 5.13 & 0.18 \\
 \rule{0pt}{0.3cm}
 & MonoUNI \cite{monouni} & None & 10.98 & 26.63 & 4.04 & 0.57 \\
 \rule{0pt}{0.3cm}
 & MonoDGP \cite{monodgp} & None & \underline{12.36} & \underline{31.12} & \underline{5.78} & \underline{1.24} \\
 \rule{0pt}{0.3cm}
 & \textbf{MonoDGP + SPAN} & None & \textbf{12.48} & \textbf{31.76} & \textbf{5.89} & \textbf{1.27} \\
\hline
\rule{0pt}{0.3cm}
\multirow{7}*{\makecell{Level\_2 \\ ($\mathrm{IoU}_{3D}\ge0.5$)}} & CaDDN \cite{caddn} & LIDAR & 16.51 & 44.87 & 8.99 & 0.58 \\
\rule{0pt}{0.3cm}
 & PCT \cite{pct} & Depth & 4.03 & 14.67 & 1.74 & 0.36 \\
 \rule{0pt}{0.3cm}
 & GUPNet \cite{gupnet} in \cite{deviant} & None & 9.39 & 24.69 & 4.67 & 0.19 \\
 \rule{0pt}{0.3cm}
 & DEVIANT \cite{deviant} & None & 10.29 & 26.75 & 4.95 & 0.16 \\
 \rule{0pt}{0.3cm}
 & MonoUNI \cite{monouni} & None & 10.38 & 26.57 & 3.95 & 0.53 \\
 \rule{0pt}{0.3cm}
 & MonoDGP \cite{monodgp} & None & \underline{11.71} & \underline{31.02} & \underline{5.61} & \underline{1.17} \\
 \rule{0pt}{0.3cm}
 & \textbf{MonoDGP + SPAN} & None & \textbf{11.95} & \textbf{31.33} & \textbf{5.78} & \textbf{1.20} \\
\bottomrule[0.15em]
\end{tabular}
\caption{Comparisons for the Vehicle Category on the Waymo val set. The best results are highlighted in \textbf{bold}, the second-best are \underline{underlined}, compared with methods without extra data.}
\vspace{-0.6em}
\label{tab:Waymo Result}
\end{table*}

The Waymo Open Dataset \cite{waymo} is a widely used benchmark for autonomous-driving scene understanding. 
It encompasses a diverse collection of real-world driving scenes and categorizes objects into LEVEL\_1 and LEVEL\_2 according to LiDAR point density.
The dataset comprises 798 training sequences and 202 validation sequences, yielding approximately 160,000 and 40,000 samples, respectively. For fair comparison, we follow the protocol of \cite{chen20153d} and subsample every third frame from both training and validation sequences, resulting in 52,386 training images and 39,848 validation images.
During evaluation, we adopt the same settings as DEVIANT \cite{deviant}. The ${AP}_{3D}$ metric is reported under two IoU thresholds (0.5 and 0.7) across four distance ranges: Overall, 0 - 30m, 30 - 50m, and 50m - $\infty $.

As shown in Table \ref{tab:Waymo Result}, our method achieves state-of-the-art performance across all ranges without the use of additional data. These results further demonstrate the effectiveness and generalizability of the proposed SPAN. Notably, the approach utilizing LiDAR outperforms SPAN. We attribute this discrepancy to the higher density of the LiDAR point clouds provided by Waymo, which better compensates for geometric ambiguities, thus leading to a greater performance gain.

\section{Compared with Prior Geometric Methods}
While sharing geometric roots, SPAN innovates by formulating consistency as a differentiable, single-stage optimization objective, differing from prior works~\cite{deep3dbox, shift_rcnn}.
The results of the comparative experiment are shown in Tab.~\ref{tab:novelty}.

\boldparagraph{Soft Constraints vs.~Hard Solving.}
~Deep3DBox~\cite{deep3dbox} uses rigid algebraic solving, where 2D noise propagates to catastrophic 3D failure (-0.81\%). SPAN's differentiable soft losses enable the network to learn resilience to 2D jitter, ensuring robustness.

\boldparagraph{Gradient Refinement vs.~Active Regression.}
~Shift R-CNN~\cite{shift_rcnn} relies on heavy multi-stage Active Regression (+15ms). SPAN achieves refinement via end-to-end gradient optimization, yielding geometric gains in a single-stage pipeline at zero inference cost.

\begin{table*}[h]
\centering
\footnotesize
\setlength{\tabcolsep}{3pt}
\renewcommand{\arraystretch}{1.0}
\begin{tabular}{l|l|c|c}
\toprule[0.15em]
Geometric Strategy & Mechanism & \makecell{Inference Cost \\[-0.3em] (Batch Size of 1)} & $AP_{3D}$ (Mod.) \\
\midrule
Baseline (MonoDGP~\cite{monodgp}) & Pure Regression & 42ms (single RTX 3090) & 22.34 \\
+ Deep3DBox~\cite{deep3dbox} & Hard Algebraic Solving & +5ms & 21.53 \textcolor{tablered}{(-0.81)} \\
+ Shift R-CNN~\cite{shift_rcnn} & Multi-stage Active Reg. & +15ms & 22.85 \textcolor{tableblue}{(+0.51)} \\
\textbf{+ SPAN (Ours)} & \textbf{Soft Loss + HTL} & \textbf{+0ms} & \textbf{23.26} \textcolor{tableblue}{\textbf{(+0.92)}} \\
\bottomrule[0.15em]
\end{tabular}
\caption{Comparison of Geometric Strategies on KITTI val set.}
\label{tab:novelty}
\vspace{-1.0em}
\end{table*}

\section{Why MGIoU?}
To address questions on MGIoU~\cite{mgiou} choice and whether gains are solely from HTL, we conducted a comprehensive ablation. As shown in Tab.~\ref{tab:ablation_detailed}.
MGIoU provides superior optimization dynamics compared to baselines. (i) Unlike L1 constraint which treats corners independently, MGIoU enforces structural volumetric integrity. (ii) Unlike Exact IoU, MGIoU utilizes projection-based approximation to provide non-vanishing gradients for disjoint boxes, ensuring superior convergence.
MGIoU outperforms L1 (+0.21\%) and Exact IoU (+0.14\%). By rigidly coupling depth and dimensions, it effectively mitigates depth bias by preventing the network from exploiting 2D projection ambiguities.

\begin{table}[h]
\centering
\small
\setlength{\tabcolsep}{5pt}
\renewcommand{\arraystretch}{1.0}
\resizebox{1.0\linewidth}{!}{
\begin{tabular}{l|c|c|c}
\toprule[0.15em]
Method Config & HTL & $AP_{3D}$ (Mod.) & $\Delta$ \\
\midrule
Baseline (MonoDGP) & - & 22.34 & - \\
Baseline + HTL Only & \checkmark & 22.56 & +0.22 \\
\midrule
Baseline + HTL + $\mathcal{L}_{corner}$ (L1) & \checkmark & 22.68 & +0.34 \\
Baseline + HTL + $\mathcal{L}_{corner}$ (Exact 3D IoU) & \checkmark & 22.75 & +0.41 \\
\textbf{Baseline + HTL + $\mathcal{L}_{corner}$ (MGIoU)} & \checkmark & \textbf{22.89} & \textbf{+0.55} \\
\bottomrule[0.15em]
\end{tabular}
}
\caption{MGIoU performance analysis.}
\label{tab:ablation_detailed}
\vspace{-0.8em}
\end{table}

\section{Disentangling HTL \& Geometric Alignment Gains}
As shown in Tab.~\ref{tab:htl_vs_loss}, we conducted an experiment replacing HTL with a simple Linear Weight Schedule to isolate gains.
By using a naive Linear Weight Schedule to mitigate initial noise, SPAN achieves 22.95\% AP, outperforming the HTL baseline. This confirms that gains stem primarily from the geometric losses.
HTL further boosts performance (+0.31\%) by adapting to training dynamics, acting as an amplifier for SPAN.

\begin{table}[h]
\centering
\footnotesize
\setlength{\tabcolsep}{3pt}
\renewcommand{\arraystretch}{1.0}
\resizebox{0.95\linewidth}{!}{
\begin{tabular}{l|c|c|c}
\toprule[0.15em]
Method Config & Schedule & $AP_{3D}$ (Mod.) & $\Delta$ \\
\midrule
Baseline & None & 22.34 & - \\
Baseline + SPAN & None & 21.85 & \textcolor{tablered}{-0.49} \\
Baseline & HTL & 22.56 & \textcolor{tableblue}{+0.22} \\
\textbf{Baseline + SPAN} & \textbf{Linear Weight Schedule} & \textbf{22.95} & \textbf{\textcolor{tableblue}{+0.61}} \\
\textbf{Baseline + SPAN} & \textbf{HTL} & \textbf{23.26} & \textbf{\textcolor{tableblue}{+0.92}} \\
\bottomrule[0.15em]
\end{tabular}
}
\caption{Decoupling Schedule vs.~Geometric Gains.}
\label{tab:htl_vs_loss}
\vspace{-0.8em}
\end{table}

\section{Interplay between Projection Alignment \& Depth Bias}
We analyze the interaction between Projection Alignment (PA) and depth bias.

\boldparagraph{PA mitigates Depth Bias.} To address the inherent scale ambiguity ($z \propto f \cdot H_{3d}/h_{2d}$), PA utilizes reliable 2D boxes as geometric anchors, rectifying depth errors that pure regression permits. Tab.~\ref{tab:depth_analysis} validates that PA reduces long-range Depth MAE, directly contributing to $AP_{3D}$ gains.

\boldparagraph{Depth Loss with Geometry Prior regularizes PA.} Conversely, depth bias loss by MonoDGP~\cite{monodgp} prevents PA from converging to degenerate solutions. By penalizing projection violations, it ensures the geometric optimization remains physically consistent.

\begin{table}[h]
\centering
\scriptsize
\setlength{\tabcolsep}{2pt}
\renewcommand{\arraystretch}{1.0}
\begin{tabular}{l|ccc|c}
\toprule[0.15em]
\multirow{2}{*}{Method} & \multicolumn{3}{c|}{Depth MAE (m) $\downarrow$ by Distance} & \multirow{2}{*}{$AP_{3D}$ (Mod.)} \\
& 0-20m & 20-40m & 40m-$\infty$ & \\
\midrule
MonoDGP & \textbf{0.34} & 0.96 & 1.55 & 22.34 \\
\textbf{+ Proj. Align + HTL} & 0.36 \textcolor{tablered}{(+0.02)} & \textbf{0.92} \textcolor{tableblue}{(-0.04)} & \textbf{1.50} \textcolor{tableblue}{(-0.05)} & \textbf{22.97} \\
\bottomrule[0.15em]
\end{tabular}
\caption{Impact of Projection Alignment on Depth Bias.}
\label{tab:depth_analysis}
\vspace{-0.8em}
\end{table}

\section{More Qualitative Results}
\begin{figure*}
  \centering
  \subfloat[ MonoDETR]
  {\includegraphics[width=0.33\linewidth]{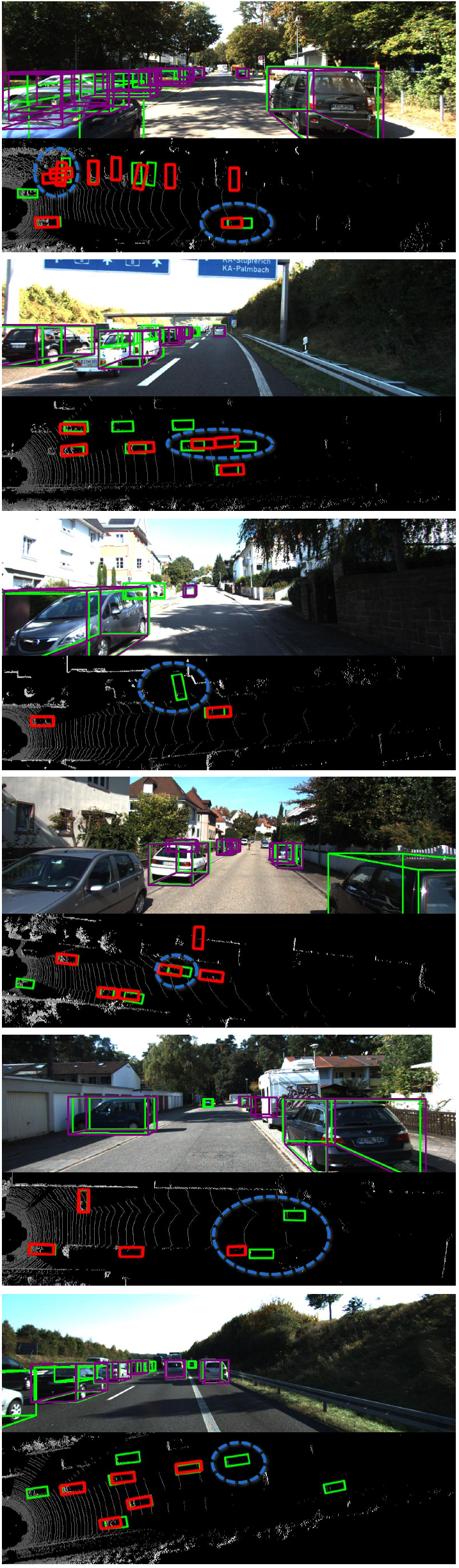}} \hfill
  \subfloat[ MonoDGP]
  {\includegraphics[width=0.33\linewidth]{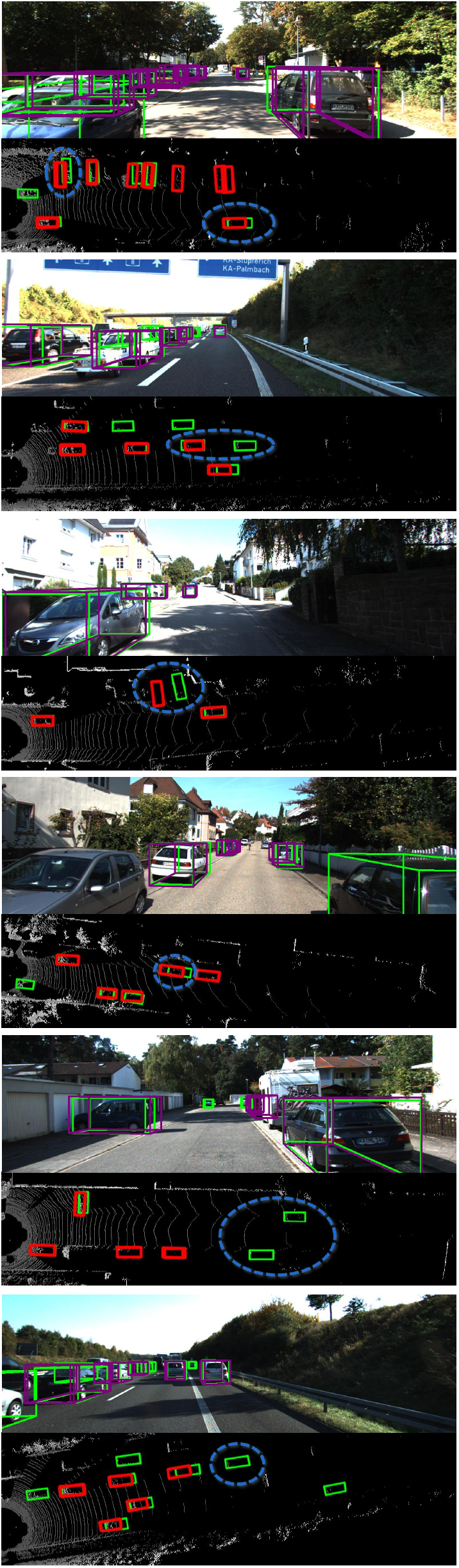}} \hfill
  \subfloat[ MonoDGP (+SPAN)]{\includegraphics[width=0.33\linewidth]{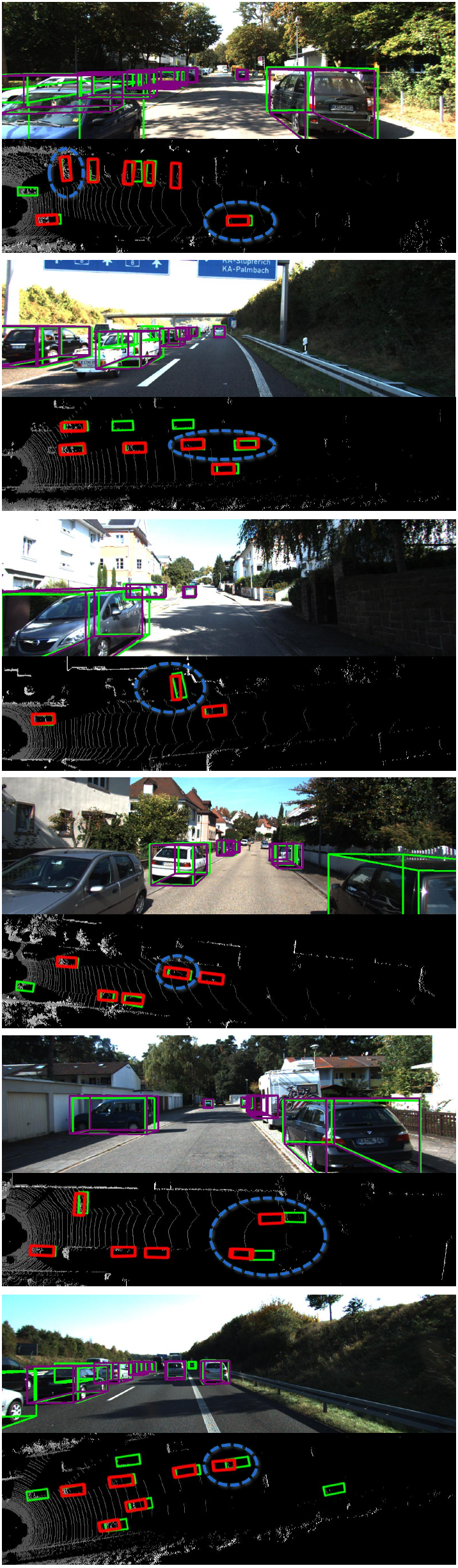}} \hfill
  \caption{\textbf{More qualitative results on KITTI val set.} (a) MonoDETR~\cite{monodetr}, (b) MonoDGP~\cite{monodgp}, and (c) MonoDGP (+SPAN). For each image set, the top row presents the camera-view visualization, while the bottom row offers the corresponding bird’s-eye view. Ground-truth bounding boxes are rendered in \textbf{\textcolor[RGB]{52,164,133}{green}}, and predictions are shown in \textbf{\textcolor[RGB]{233,69,32}{red}}. We also circle some objects to highlight the difference between the baseline model and our method.}
  \label{fig:more qualitative res}
\end{figure*}

Additional qualitative comparisons on the KITTI validation set are presented in Fig.~\ref{fig:more qualitative res}. The proposed method consistently delivers more accurate 3D localization, especially for distant and small objects, thereby demonstrating its superior robustness under challenging scale variations.

\end{document}